\documentclass{article}

\usepackage{microtype}
\usepackage{graphicx}
\usepackage{subfigure}
\usepackage{booktabs} % for professional tables
\usepackage{colortbl}
\usepackage{makecell}

\usepackage{hyperref}

\usepackage[accepted]{icml2023}

\usepackage{amsmath}
\usepackage{amssymb}
\usepackage{mathtools}
\usepackage{amsthm}
\usepackage{float}

\usepackage[capitalize,noabbrev]{cleveref}

\theoremstyle{plain}

\theoremstyle{definition}

\theoremstyle{remark}

\usepackage[textsize=tiny]{todonotes}

\usepackage{amsmath,amsfonts,bm}
\usepackage[most]{tcolorbox}

\def\1{\bm{1}}

\def\vzero{{\bm{0}}}

\def\vmu{{\bm{\mu}}}
\def\vtheta{{\bm{\theta}}}
\def\vphi{{\bm{\phi}}}

\def\vepsilon{{\bm{\epsilon}}}

\def\vx{{\bm{x}}}
\def\vy{{\bm{y}}}
\def\vz{{\bm{z}}}

\def\mI{{\bm{I}}}

\DeclareMathAlphabet{\mathsfit}{\encodingdefault}{\sfdefault}{m}{sl}
\SetMathAlphabet{\mathsfit}{bold}{\encodingdefault}{\sfdefault}{bx}{n}

\def\gD{{\mathcal{D}}}
\def\gE{{\mathcal{E}}}

\def\gN{{\mathcal{N}}}

\newcommand{\E}{\mathbb{E}}

\usepackage[most]{tcolorbox}

\newtcolorbox{mblock}[1]
{
  colframe     = gray,
  coltitle     = black,
  colbacktitle = lightgray!50!white,
  colback      = white,
  title        = #1,
  fonttitle    = \bfseries,
  arc          = 0mm,
  left         = 2pt,
  right        = 2pt,
}

\newtcolorbox{rblock}[1]
{
  colframe     = green,
  coltitle     = gray,
  colbacktitle = lightgray!50!white,
  colback      = white,
  title        = \ifthenelse{\isempty{#1}}
                  {Remark}
                  {Remark: #1},
  fonttitle    = \bfseries,
  arc          = 0mm,
  left         = 2pt,
  right        = 2pt,
}

\def\ours{{UniDiffuser}}

\def\txt{{\mathrm{text}}}
\icmltitlerunning{One Transformer Fits All Distributions in Multi-Modal Diffusion at Scale}

\begin{document}

\twocolumn[
\icmltitle{One Transformer Fits All Distributions in Multi-Modal Diffusion at Scale}



\icmlsetsymbol{equal}{*}

\begin{icmlauthorlist}
\icmlauthor{Fan Bao}{thu,ss}
\icmlauthor{Shen Nie}{renmin,ss}
\icmlauthor{Kaiwen Xue}{renmin,ss}
\icmlauthor{Chongxuan Li}{renmin}
\icmlauthor{Shi Pu}{thu,ss}
\icmlauthor{Yaole Wang}{thu}\\
\icmlauthor{Gang Yue}{thu,ss}
\icmlauthor{Yue Cao}{zhiyuan}
\icmlauthor{Hang Su}{thu,lab}
\icmlauthor{Jun Zhu}{thu,ss,lab}
\end{icmlauthorlist}

\icmlaffiliation{thu}{Dept. of Comp. Sci. \& Tech., Institute for AI, Tsinghua-Huawei Joint Center for AI, BNRist Center, State Key Lab for Intell. Tech. \& Sys., Tsinghua University}
\icmlaffiliation{renmin}{Gaoling School of AI, Renmin University of China; Beijing Key Lab of Big Data Management and Analysis Methods, Beijing, China}
\icmlaffiliation{zhiyuan}{Beijing Academy of Artificial Intelligence}
\icmlaffiliation{ss}{ShengShu, Beijing, China}
\icmlaffiliation{lab}{Pazhou Laboratory (Huangpu), Guangzhou, China}

\icmlcorrespondingauthor{Jun Zhu}{dcszj@tsinghua.edu.cn}

\icmlkeywords{Machine Learning, ICML}

\vskip 0.3in
]



\printAffiliationsAndNotice{}  % leave blank if no need to mention equal contribution

\begin{abstract}
This paper proposes a unified diffusion framework (dubbed \emph{\ours{}}) to \emph{fit all distributions 
relevant to a set of multi-modal data in one model}.
Our key insight is -- learning diffusion models for marginal, conditional, and joint distributions can be unified as predicting the noise in the perturbed data, where the perturbation levels (i.e. timesteps) can be different for different modalities. Inspired by the unified view, \ours{} learns all distributions simultaneously with a minimal modification to the original diffusion model -- perturbs data in all modalities instead of a single modality, inputs individual timesteps in different modalities, and predicts the noise of all modalities instead of a single modality. \ours{} is parameterized by a transformer for diffusion models to handle input types of different modalities.
Implemented on large-scale paired image-text data, 
\ours{} is able to perform image, text, text-to-image, image-to-text, and image-text pair generation by setting proper timesteps without additional overhead. In particular, \ours{} is able to produce perceptually realistic samples in all tasks and its quantitative results (e.g., the FID and CLIP score) are not only superior to existing general-purpose models but also comparable to the bespoken models (e.g., Stable Diffusion and DALL$\cdot$E 2) in representative tasks (e.g., text-to-image generation). Our code is available at \url{https://github.com/thu-ml/unidiffuser}.
\end{abstract}

\section{Introduction}

% TODO. Recently, generative modeling on multi-modal data has garnered significant attention and has been one of the most promising ways to synthesize data. 
% Indeed, we are witnessing a content-creation revolution driven by the rapid advances of generative modeling.

Recently, we are witnessing a content-creation revolution driven by the rapid advances of generative modeling on multi-modal data.
In particular, diffusion models~\cite{sohl2015deep,ho2020denoising,song2020score} have shown an incredible ability to create high-fidelity and diverse data~\cite{ramesh2022hierarchical,saharia2022photorealistic,rombach2022high,ho2022imagen,popov2021grad},
% , including images~\cite{ramesh2022hierarchical,saharia2022photorealistic,rombach2022high}, videos~\cite{ho2022imagen}, and audios~\cite{popov2021grad}, 
whose content aligns well with the input text condition. 

\begin{figure*}[t]
\centering
\includegraphics[width=\linewidth]{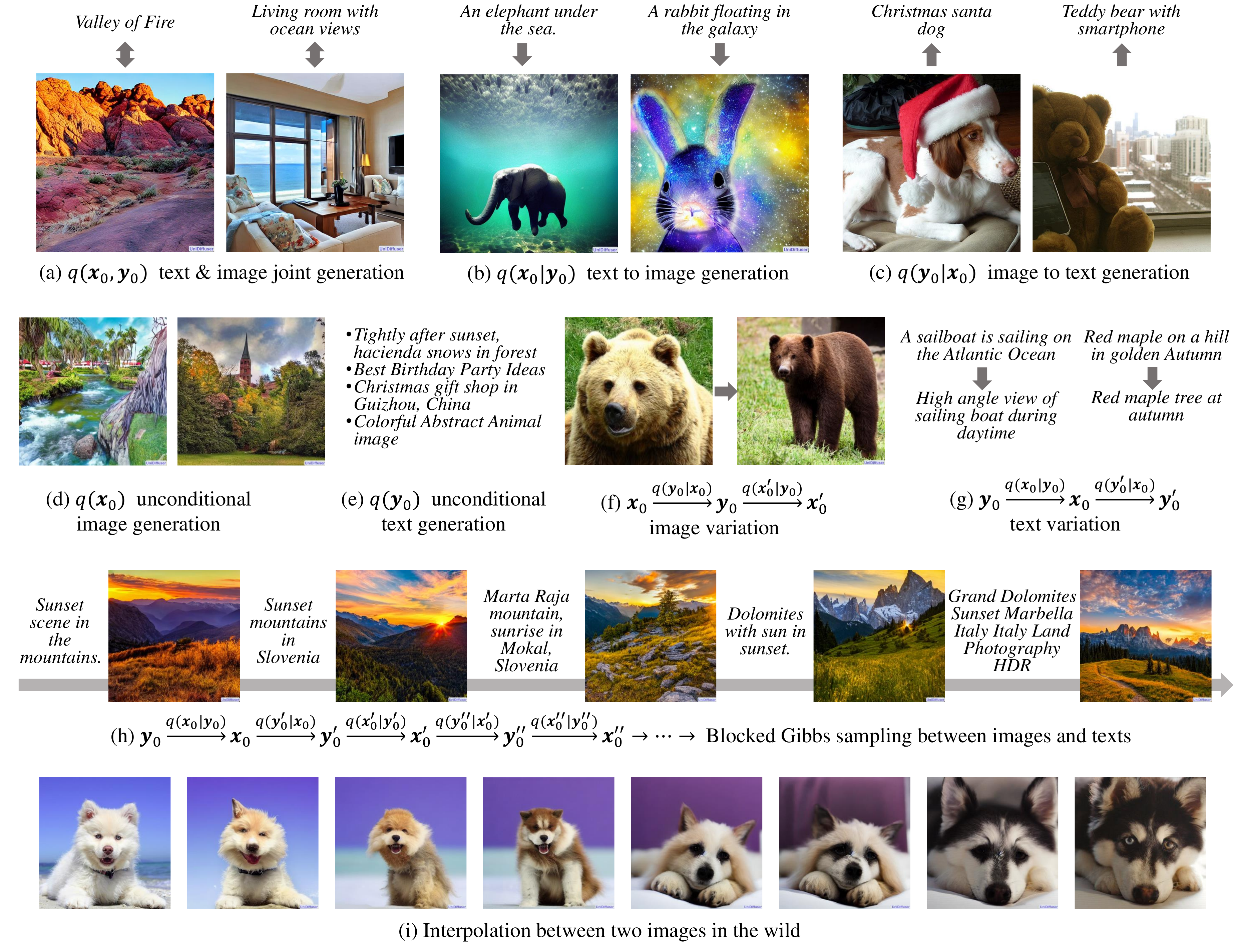}
\caption{\textbf{\ours{} handles various tasks by fitting all distributions with one transformer.} (a-e) \ours{} can directly perform joint generation, conditional generation, and unconditional generation. (f-g) Image variation and text variation are direct applications by leveraging two conditional distributions modeled by \ours{}. (h) Furthermore, \ours{} can perform blocked Gibbs sampling to see how images and texts are translated to each other. (i) \ours{} can also perform interpolation between two images in the wild.}
\label{fig:demos}
\end{figure*}

However, these generative models are designed as bespoke systems, which only allow a single task.
Actually, humans can generate various multi-modal content simultaneously, with arbitrary conditioning types. 
For example, artists can create paintings conditioned on texts, scenes, or just imagination and employ language ability to generate the caption of a photo. 
Toward a general generative system on multi-modal data, a unified training framework that can cover all types of multi-modal generative tasks (see Figure~\ref{fig:demos}) 
% \yue{no definition on ``all types of tasks": that can cover all types of multi-modal generative tasks} 
is one of the fundamental components.

% \yue{However, these generative models are designed as bespoke systems, which only allow a single generation task, e.g., image generation, text-to-image generation, and image captioning. Actually, humans can generate various multi-modal content simultaneously, with arbitrary conditioning types. As an example, artists can create paintings using texts, images, or just imagination as conditioning and employ language capability to generate the caption of an image. 
% Toward a general generative system on multi-modal data, there are two fundamental components -- the architecture that can handle input types of different modalities, and the training framework which can cover all types of generative tasks. The emergence of Transformer~\cite{Attn,ViT} and its application on generative modeling~\cite{UViT} provide the foundation architecture for the general generative system. This paper presents a unified probabilistic modeling framework with diffusion model (dubbed \emph{\ours{}}) to capture the joint, conditional, and marginal distributions simultaneously (see Figure~\ref{fig:demos}).}

The task is solved by fitting a corresponding distribution in the view of probabilistic modeling. For instance, text-to-image generation can be formulated as learning the conditional distribution $p(\textrm{Image}|\textrm{Text})$. 
A classical way to fit all relevant distributions is implicit -- it first learns the joint distribution and then infers the marginal and conditional distributions by additional procedures (e.g., Markov Chain Monte Carlo~\cite{srivastava2012multimodal}), which is unaffordable on large-scale multi-modal data~\cite{schuhmann2022laion}.

\begin{figure*}[t]
\centering
\includegraphics[width=\linewidth]{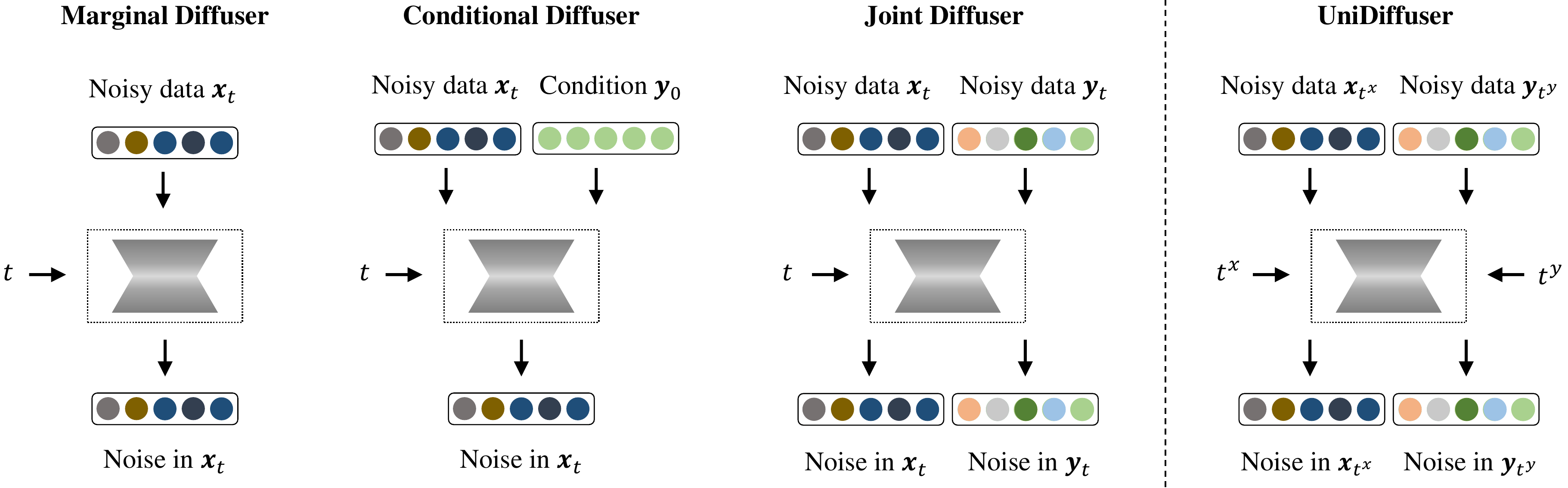}
\caption{\textbf{Comparison with bespoken diffusers.} \ours{} fits all distributions simultaneously with a minimal modification of \citet{ho2020denoising}. In particular, it degenerates to bespoken diffusion models by setting the timesteps (or noise levels) properly.}
\label{fig:compare_with_ddpm}
\end{figure*}

In contrast, this paper presents a diffusion-based framework (dubbed \emph{\ours{}}) that \emph{explicitly fits all relevant distributions in one model} without introducing additional training or inference overhead. 
Our key insight is -- learning diffusion models for all distributions can be unified as predicting the noise in the perturbed data, where the perturbation levels (i.e. timesteps) can be different for different modalities. For instance, a zero level indicates conditional generation given the corresponding modality, and a maximum level indicates unconditional generation of other modalities by ignoring the corresponding modality. Inspired by the unified view, \ours{} learns all distributions simultaneously with a minimal modification to the original diffusion model~\cite{ho2020denoising} (see Figure~\ref{fig:compare_with_ddpm}) -- perturbs data in all modalities instead of a single modality, inputs individual timesteps in different modalities, and predicts the noise of all modalities instead of a single modality. 
Naturally, \ours{} is able to perform all kinds of generation (see Figure~\ref{fig:demos}) in the same way as bespoken diffusion models.
Moreover, \ours{} can perform the classifier-free guidance~\cite{ho2021classifier} for free to improve the sample quality in both conditional and joint generation because \ours{} already models marginal distributions.

Besides the probabilistic modeling framework, a unified architecture that can handle input types of different modalities is another fundamental component in a general generative system. Notably, the emergence of Transformer~\cite{vaswani2017attention,dosovitskiy2020image} and its applications on generative modeling~\cite{bao2022all} provide a promising solution to capture interactions between modalities. Naturally, \ours{} employs a transformer-based backbone.

We implement \ours{} in the latent space~\cite{rombach2022high} with an additional CLIP encoder~\cite{radford2021learning} for images and GPT-2~\cite{radford2019language} decoder for texts on large-scale image-text data~\cite{schuhmann2022laion}.
\ours{} is able to perform image, text, text-to-image, image-to-text, and image-text pair generation by setting proper timesteps without additional overhead. In particular, \ours{} is able to produce perceptually realistic samples in all tasks and its quantitative results (e.g., the FID and CLIP score) are not only superior to existing general-purpose models but also comparable to the corresponding bespoken models (e.g., Stable Diffusion and DALL$\cdot$E 2) in representative tasks (e.g., text-to-image generation).

% We firstly convert images and texts to latent embeddings via an image encoder and a text encoder, and then use \ours{} to model these latent embeddings. Specifically, we adopt U-ViT~\cite{bao2022all}, a transformer-based architecture, as the backbone of the joint noise prediction network. U-ViT treats all inputs, including latent embeddings and the timesteps for different modalities, as tokens. To improve the numerical stability of large U-ViT, we adjust the position of the layer normalization, thereby enabling stable training with mixed precision. 

% Contributions:
% \begin{itemize}
%     \item One model for all distributions
%     \item Only one objective, one feedforward is able to compute. Other multi-modal works (VD) have multiple objectives, need to feedforward multiple times.
%     \item We demonstrate the possibility to employ transformers in large-scale diffusion models, which is not limited to multi-modal diffusion, but also prior text-to-image diffusion models
%     \item We are the first to train large diffusion models with transformers
%     \item The backbone is simple. Only use one network. VD use two network.
%     \item Transformers' motivation: big convergence
%     \item Validate that a single weight can tackle both the image and text information, in the area of generative modeling
% \end{itemize}

% \begin{enumerate}
%     \item Universal Algorithm for training and inference -> MTDM.
%     \item Universal Model -> transformer
%     \item Experiments at scale.
% \end{enumerate}

\section{Background}
\label{sec:background}

\textbf{Diffusion models}~\cite{sohl2015deep,ho2020denoising} perturb the data by gradually injecting noise to data $\vx_0 \sim q(\vx_0)$, which is formalized by a Markov chain:
\begin{align*}
    & q(\vx_{1:T}|\vx_0) = \prod\limits_{t=1}^T q(\vx_t|\vx_{t-1}),\\
    & q(\vx_t|\vx_{t-1}) = \gN(\vx_t|\sqrt{\alpha_t} \vx_{t-1}, \beta_t \mI),
\end{align*}
where $\beta_t$ is the noise schedule and $\alpha_t = 1 - \beta_t$.

The data can be generated by reversing this process, where the reverse transition $q(\vx_{t-1}|\vx_t)$ is approximated by a Gaussian model $p(\vx_{t-1}|\vx_t) = \gN(\vx_{t-1}|\vmu_t(\vx_t), \sigma_t^2 \mI)$. As shown by \citet{bao2022analytic}, the optimal mean under maximal likelihood estimation is
\begin{align}
\label{eq:opt_mean}
    \vmu_t^*(\vx_t) = \frac{1}{\sqrt{\alpha_t}} \left(\vx_t - \frac{\beta_t}{\sqrt{1-\overline{\alpha}_t}} \E[\vepsilon^x|\vx_t]\right),
\end{align}
where $\overline{\alpha}_t = \prod_{i=1}^t \alpha_i$ and $\vepsilon^x$ is the standard Gaussian noise injected to $\vx_t$. To estimate the conditional expectation $\E[\vepsilon^x|\vx_t]$, a noise prediction network $\vepsilon_\vtheta(\vx_t, t)$ is adopted to minimize the regression loss as follows:
\begin{align}
\label{eq:l2}
    \min\limits_\vtheta \E_{t, \vx_0, \vepsilon^x} \| \vepsilon^x - \vepsilon_\vtheta(\vx_t, t) \|_2^2,
\end{align}
where $t$ is uniformly sampled from $\{1,2,\dots,T\}$ and $\vx_t = \sqrt{\overline{\alpha}_t} \vx_0 + \sqrt{1 - \overline{\alpha}_t} \vepsilon^x$. According to the property of the $l_2$ regression loss, the optimal noise prediction network satisfies $\vepsilon_{\vtheta^*}(\vx_t, t) = \E[\vepsilon^x|\vx_t]$. Since Eq.~{(\ref{eq:l2})} is also equivalent to the denoising score matching loss~\cite{vincent2011connection}, the optimal noise prediction network also satisfies $\vepsilon_{\vtheta^*}(\vx_t, t) = - \sqrt{\overline{\beta}_t} \nabla \log q(\vx_t)$, where $q(\vx_t)$ is the distribution of the perturbed data at timestep $t$.

\textbf{Conditional generation with diffusion models.} In the case of conditional generation, we have paired data $(\vx_0, \vy_0) \sim q(\vx_0, \vy_0)$, and we want to model the conditional data distribution $q(\vx_0|\vy_0)$. The Gaussian model of the reverse process conditioned on $\vy_0$ is $p(\vx_{t-1}|\vx_t, \vy_0) = \gN(\vx_{t-1}|\vmu_t(\vx_t, \vy_0), \sigma_t^2\mI)$. Similarly to Eq.~{(\ref{eq:opt_mean})}, the optimal mean under maximal likelihood estimation is
\begin{align}
\label{eq:opt_mean_c}
    \vmu_t^*(\vx_t, \vy_0) = \frac{1}{\sqrt{\alpha_t}} \left(\vx_t - \frac{\beta_t}{\sqrt{1-\overline{\alpha}_t}} \E[\vepsilon^x|\vx_t, \vy_0]\right).
\end{align}
To estimate $\E[\vepsilon^x|\vx_t, \vy_0]$, a noise prediction network conditioned on $\vy_0$ is adopted to minimize the regression loss
\begin{align*}
    \min\limits_\vtheta \E_{t, \vx_0, \vy_0, \vepsilon^x} \| \vepsilon^x - \vepsilon_\vtheta(\vx_t, \vy_0, t) \|_2^2.
\end{align*}
% Such a conditional diffusion model is able to generate one modality given another in a single direction.

\textbf{Classifier-free guidance (CFG)}~\cite{ho2021classifier} is proposed to improve the sample quality of a conditional diffusion model. Specifically, it samples by linearly combining a conditional model and an unconditional one:
\begin{align}
\label{eq:cfg}
    \hat{\vepsilon}_\vtheta(\vx_t, \vy_0, t) = (1 + s) \vepsilon_\vtheta(\vx_t, \vy_0, t) - s \vepsilon_\vtheta(\vx_t, t),
\end{align}
where $s$ is the guidance scale.
The conditional and unconditional models share parameters by introducing a null token $\varnothing$, i.e., $\vepsilon_\vtheta(\vx_t, t) = \vepsilon_\vtheta(\vx_t, \vy_0=\varnothing, t)$.

\section{Method}

Section~\ref{sec:train} presents \emph{\ours{}}, a single diffusion model to capture the marginal, conditional, and joint distributions determined by multi-modal data simultaneously. Section~\ref{sec:cfg_free} demonstrates how to perform classifier-free guidance (CFG) for free in conditional and joint sampling of \ours{}. For simplicity, we focus on two-modal data in this paper but \ours{} can be easily extended to more modalities. 

\subsection{\ours{}: One Diffusion Fits All Distributions}
\label{sec:train}
Formally, suppose we have two modalities of data sampled from distribution $q(\vx_0, \vy_0)$. We aim to design a diffusion-based model that is able to capture all relevant distributions determined by $q(\vx_0, \vy_0)$, i.e., the marginal distributions $q(\vx_0)$ and $q(\vy_0)$, the conditional distributions $q(\vx_0|\vy_0)$ and $q(\vy_0|\vx_0)$, and the joint distribution $q(\vx_0, \vy_0)$.

We notice that learning a distribution with diffusion models is equivalent to estimating a conditional expectation over the noise. In particular, modeling the marginal distribution $q(\vx_0)$ is equivalent to estimating the conditional expectation of the noise injected to $\vx_t$, i.e., $\E[\vepsilon^x|\vx_t]$, according to Eq.~{(\ref{eq:opt_mean})}.
Similarly, the key quantities to be estimated in modeling the conditional distribution $q(\vx_0|\vy_0)$  and the joint distribution $q(\vx_0, \vy_0)$ are $\E[\vepsilon^x|\vx_t, \vy_0]$ (see Eq.~{(\ref{eq:opt_mean_c})}) and $\E[\vepsilon^x, \vepsilon^y|\vx_t, \vy_t]$ respectively.

% According to Eq.~{(\ref{eq:opt_mean})}, the key to model the marginal distributions $q(\vx_0)$ and $q(\vy_0)$ is to estimate the conditional expectation of the noise injected to $\vx_t$ and $\vy_t$, i.e., $\E[\vepsilon^x|\vx_t]$ and $\E[\vepsilon^y|\vy_t]$ respectively. According to Eq.~{(\ref{eq:opt_mean_c})}, the key to model the conditional distributions $q(\vx_0|\vy_0)$ and $q(\vy_0|\vx_0)$ is to estimate $\E[\vepsilon^x|\vx_t, \vy_0]$ and $\E[\vepsilon^y|\vy_t, \vx_0]$ respectively. Similarly, the key to model the joint distribution $q(\vx_0, \vy_0)$ is to estimate $\E[\vepsilon^x, \vepsilon^y|\vx_t, \vy_t]$.

% Our key observation is that learning a distributions with diffusion models is equivalent to estimating a conditional expectation over the noise, whose condition consists of
% perturbed data with potentially different noise levels in all modalities.

% We unify learning different distributions with diffusion models as estimating a conditional expectation over the noise, whose condition consists of
% perturbed data with potentially different noise levels in all modalities. In particular, a zero level means conditioning on the corresponding modality and a maximum level means marginalizing it. Inspired by the unified perspective, 
% \ours{} injects noise with independent timesteps for each modality and trains a \textit{joint noise prediction network} to  estimate all conditional expectations by predicting all noise together. 

A key observation is that all above conditional expectations can be unified in the general form of $\E[\vepsilon^x, \vepsilon^y|\vx_{t^x}, \vy_{t^y}]$, where $t^x$ and $t^y$ are two timesteps that can be different, 
and $\vx_{t^x}$ and $\vy_{t^y}$ are the corresponding perturbed data. 
In particular, a maximum timestep  $T$ means marginalizing it. Namely, by setting $t^{y}=T$, we have $\E[\vepsilon^x|\vx_{t^x}, \vy_T] \approx \E[\vepsilon^x|\vx_{t^x}]$\footnote{
There is a negligible gap between $\vy_T$ and the standard Gaussian noise $\vepsilon^y$ for a large $T$ (e.g., 1000 by default~\cite{ho2020denoising}).}, which corresponds to the marginal distribution $q(\vx_0)$. Similarly, a zero timestep means conditioning on the corresponding modality and a tied timestep means sampling two modalities jointly. Formally, $\E[\vepsilon^x|\vx_{t^x}, \vy_0]$  corresponds to the conditional distribution $q(\vx_0|\vy_0)$ by setting $t^y = 0$ and $\E[\vepsilon^x, \vepsilon^y|\vx_t, \vy_t]$ corresponds to the joint distribution $q(\vx_0, \vy_0)$ by setting $t^x = t^y = t$. 
Moreover, we can characterize
$q(\vx_0 | \vy_{t^y}) $ and 
$q(\vy_0 | \vx_{t^x}) $ for all $t^y$ and $t^x$ and generate data conditioned on noisy input,
by estimating $\E[\vepsilon^x, \vepsilon^y|\vx_{t^x}, \vy_{t^y}]$ in general.

% \fan{other settings.}
% \hangx{I think this is one major contribution, and we may inlcude more analysis about how we address the challenges by considering different time steps}
% \hangx{a network? maybe not a common representation?}

Inspired by the unified view, we learn $\E[\vepsilon^x, \vepsilon^y|\vx_{t^x}, \vy_{t^y}]$ for all $0\le t^x, t^y \le T$ to model all relevant distributions determined by $q(\vx_0, \vy_0)$.  Specifically, we employ a \textit{joint noise prediction network}\footnote{\ours{} can be easily reparameterized to data prediction or velocity prediction~\cite{salimans2022progressive} as well.} $\vepsilon_\vtheta (\vx_{t^x}, \vy_{t^y}, t^x, t^y)$ to predict the noise injected to $\vx_{t^x}$ and $\vy_{t^y}$ together by minimizing the following regression loss similarly to~\citet{ho2020denoising}:
\begin{align}
\label{eq:j_l2}
    & \E_{\vx_0,\vy_0, \vepsilon^x,\vepsilon^y,t^x,t^y} \|\vepsilon_\vtheta(\vx_{t^x},\vy_{t^y}, t^x,t^y) - [\vepsilon^x,\vepsilon^y] \|_2^2,
    % = & \E_{\vx_0,\vy_0, \vepsilon^x,\vepsilon^y,t^x,t^y} \left[ \|\vepsilon_\vtheta^x(\vx_{t^x},\vy_{t^y}, t^x,t^y) - \vepsilon^x \|_2^2 \right. \nonumber \\
    % & \qquad \qquad \qquad \quad \left. + \|\vepsilon_\vtheta^y(\vx_{t^x},\vy_{t^y}, t^x,t^y) - \vepsilon^y \|_2^2 \right], \nonumber
\end{align}
where $(\vx_0, \vy_0)$ is a random data point, $[,]$ denotes concatenation, $\vepsilon^x$ and $\vepsilon^y$ are sampled from standard Gaussian distributions, and $t^x$ and $t^y$ are uniformly sampled from $\{1,2,\dots,T\}$ independently. We call our method \textit{\ours{}} because it captures multiple distributions in a unified way. 
We present the training algorithm in Appendix~\ref{sec:alg}. 

% \footnote{We can also restrict $t^x$ and $t^y$ to values required during inference, but we find it does not perform better.}

The objective in Eq.~(\ref{eq:j_l2}) is as simple as the original DDPM in Eq.~{(\ref{eq:l2})}. Besides, for a single update of parameters, \ours{} only requires a single forward-backward calculation for multiple tasks (i.e., distributions), which is as efficient as the original DDPM. Although the gradient estimate of \ours{} has a slightly higher variance than the original DDPM due to two independent timesteps, we do not observe that \ours{} suffers from slower convergence.

% % since our method adopts different timescales for different modalities. 

% \cx{Training as simple and as efficient. No slow convergence.}

% \subsection{Ability for All Kinds of Sampling}

\ours{} attempts to fit all distributions by one joint noise prediction network, requiring that the backbone can handle the mutual interaction between modalities and is scalable for large-scale data and multiple tasks. Inspired by the excellent performance of transformers on multi-modal representation learning at scale~\cite{kim2021vilt,wang2022image}, we employ a transformer-based network in \ours{}, as detailed in Section~\ref{sec:uvit}.

Given a single joint noise prediction network, \ours{} can perform unconditional, conditional, and joint sampling according to a certain sampler (see Appendix~\ref{sec:alg} for the sampling algorithm). Notably, by setting the timesteps properly, the inference procedure of \ours{} is the same as the bespoken models.  In comparison, learning a single joint distribution~\cite{srivastava2012multimodal,hu2022unified} over multi-modal data requires additional procedures (e.g., Markov Chain Monte Carlo) to sample from the marginal or conditional distributions, which is unaffordable on large-scale multi-modal data~\cite{schuhmann2022laion}.

\subsection{Classifier-Free Guidance for Free}
\label{sec:cfg_free}

\begin{figure}
    \centering
    \includegraphics[width=\columnwidth]{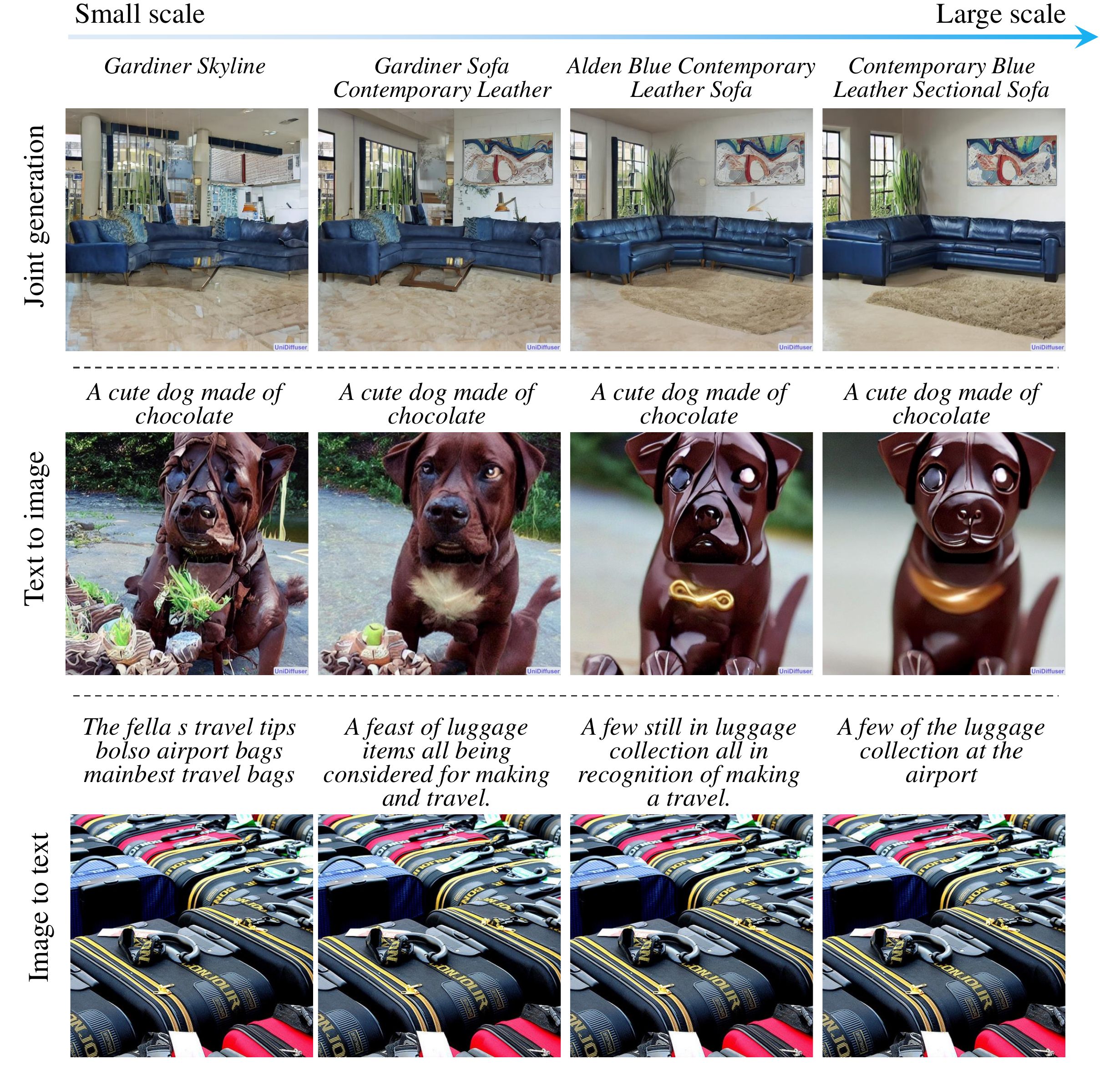}
    \caption{\textbf{Effects of CFG.} \ours{} employs CFG for free in joint and conditional sampling, improving the sample quality and image-text alignment with a large scale of around 6.}
    \label{fig:cfg}
\end{figure}

Classifier-free guidance (CFG)~\cite{ho2021classifier} combines a conditional and an unconditional model linearly during sampling (see Eq.~{(\ref{eq:cfg})}). It is simple yet effective to improve the sample quality and image-text alignment in diffusion models. Notably, CFG is directly applicable to the conditional and joint sampling of \ours{} without modifying the training process (see Figure~\ref{fig:cfg} for results).

Formally, we denote the output of $\vepsilon_\vtheta$ as the concatenation of $\vepsilon_\vtheta^x$ and $\vepsilon_\vtheta^y$, i.e. $\vepsilon_\vtheta= [\vepsilon_\vtheta^x, \vepsilon_\vtheta^y]$, where we omit the input for simplicity. \ours{} can perform CFG for free in conditional sampling because it captures both the conditional and unconditional models. For example, we can generate $\vx_0$ conditioned on $\vy_0$ similarly to Eq.~{(\ref{eq:cfg})} as follows:
\begin{align*}
\hat{\vepsilon}_\vtheta^x(\vx_t, \vy_0, t) = (1+s)\vepsilon_\vtheta^x(\vx_t, \vy_0, t, 0) -s \vepsilon_\vtheta^x(\vx_t, \vepsilon^y, t, T),
\end{align*}
where $\vepsilon_\vtheta^x(\vx_t, \vy_0, t, 0)$ and $\vepsilon_\vtheta^x(\vx_t, \vepsilon^y, t, T)$  represent the conditional and unconditional models respectively, and $s$ is the guidance scale. In contrast to the original CFG, \ours{} does not need to specify a null token for parameter sharing.

CFG is also applicable to joint sampling. By setting $t^x = t^y = t$, note that the joint score model can be equivalently expressed in the form of conditional models as follows:
\begin{align*}
     \vepsilon_\vtheta (\vx_t , \vy_t , t , t) \! \approx \! & - \! \sqrt{\overline{\beta}_t} [\nabla_{\vx_t} \! \log q(\vx_t, \vy_t),  \! \nabla_{\vy_t} \! \log q(\vx_t, \vy_t)] \\
     \! =  \! & -  \! \sqrt{\overline{\beta}_t} [\nabla_{\vx_t}  \! \log q(\vx_t | \vy_t),  \! \nabla_{\vy_t}  \! \log q(\vy_t | \vx_t)],
\end{align*}
where $q(\vx_t, \vy_t)$ is the joint distribution of perturbed data at the same noisy level $t$. 
Inspired by the above relationship between score functions,   $\vepsilon_\vtheta(\vx_t, \vy_t, t, t)$ can be viewed as approximating a pair conditional scores $\nabla_{\vx_t} \log q(\vx_t | \vy_t)$  and $\nabla_{\vy_t} \log q(\vy_t | \vx_t)$. In the same spirit of CFG, we can replace each conditional score by interpolating the joint model with the corresponding unconditional model as follows:
\begin{align*}
    & \hat{\vepsilon}_\vtheta(\vx_t, \vy_t, t) \\
     = & (1\!+\!s) \vepsilon_\vtheta(\vx_t, \vy_t, t, t)\!-\! s [\vepsilon_\vtheta^x(\vx_t, \vepsilon^y, t, T),\! \vepsilon_\vtheta^y(\vepsilon^x, \vy_t, T, t)] \\
    \approx & - \sqrt{\overline{\beta}_t} [(1+s)\nabla_{\vx_t} \log q(\vx_t | \vy_t) -s \nabla_{\vx_t} \log q(\vx_t),\\
    & \qquad \qquad (1+s)\nabla_{\vy_t} \log q(\vy_t | \vx_t) -s \nabla_{\vy_t} \log q(\vy_t)], \\
\end{align*}
where $\vepsilon_\vtheta^x(\vx_t, \vepsilon^y, t, T)$ and $\vepsilon_\vtheta^y(\vepsilon^x, \vy_t, T, t)$ represent unconditional models. We summarize the formulation of CFG in \ours{} for all tasks in Appendix~\ref{sec:summary_cfg}.
% \hangx{could be included in the main body?}

\begin{figure*}
    \centering
\includegraphics[width=\linewidth]{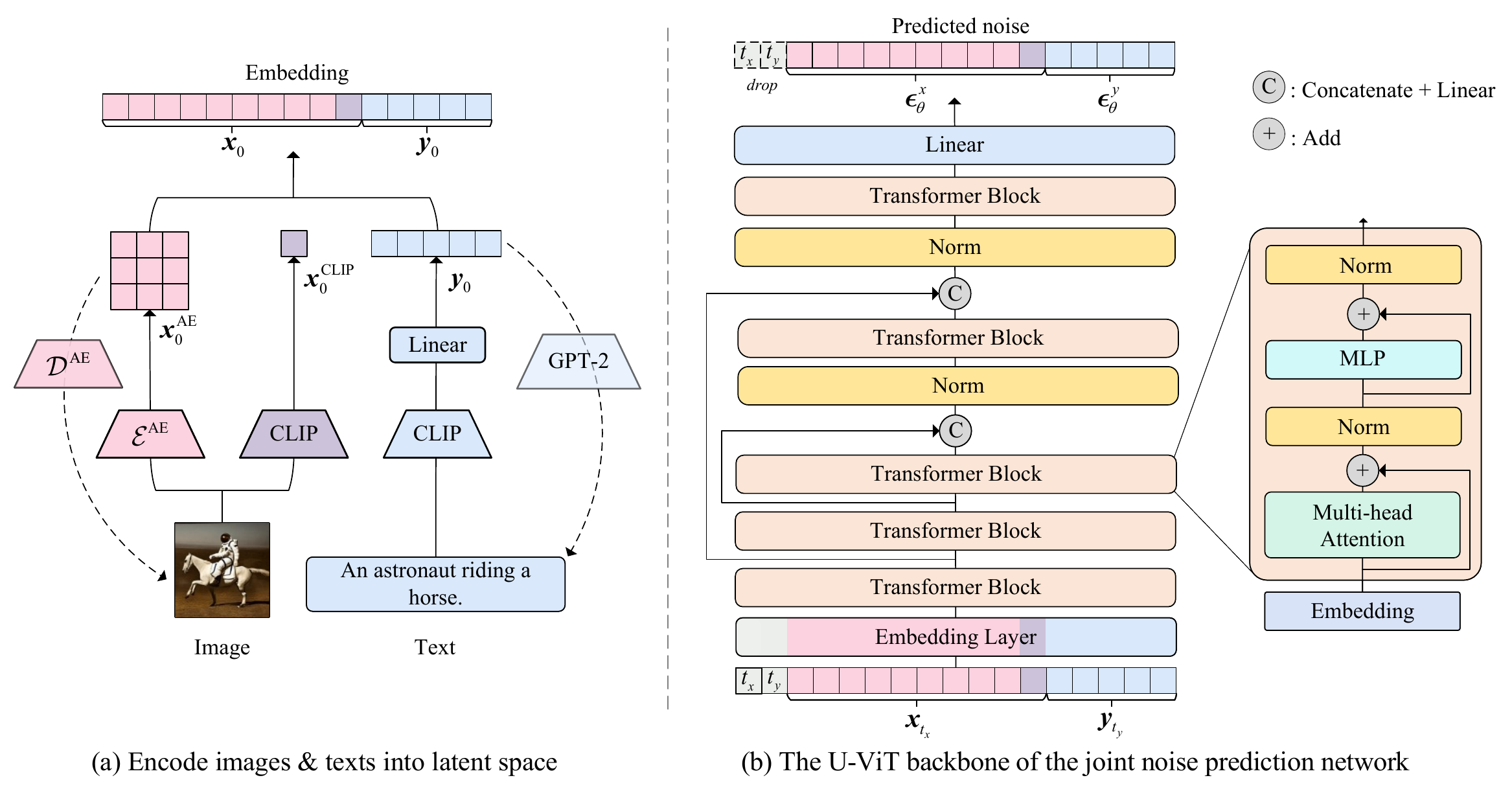}
    \caption{\textbf{Implementation of \ours{} on image-text data.} (a) First, we encode images and texts into latent space. (b) Second, we train \ours{} parameterized by a transformer~\cite{bao2022all} in the way illustrated in Figure~\ref{fig:compare_with_ddpm}  on the latent embeddings.}
    \label{fig:pipeline}
\end{figure*}

\section{\ours{} on Images and Texts}

% $\vx_0 = \gE^\img (\mathbf{I})$
% $\vy_0 = \gE^\txt (\mathbf{T})$
% $\gD^\img$
% $\gD^\txt$

Images and texts are two of the most common modalities in daily life. Thus, it is representative to validate the effectiveness of \ours{} on the two modalities.

Our implementation is two-staged  following~\cite{rombach2022high}  (see Figure~\ref{fig:pipeline}). First, we convert images and texts to continuous latent embeddings $\vx_0$ and $\vy_0$ via image and text encoders and introduce two decoders for reconstruction, as presented in Section~\ref{sec:encode}. Second, we train \ours{} parameterized by a transformer on the latent embeddings $\vx_0$ and $\vy_0$, as presented in Section~\ref{sec:uvit}.

\subsection{Encoding Images and Texts into Latent Space}
\label{sec:encode}

The image and text encoder-decoders are illustrated in Figure~{\ref{fig:pipeline} (a)}. Below we provide their details.

\textbf{Image encoder-decoder.} The image encoder consists of two parts. The first part is the image autoencoder employed in Stable Diffusion~\cite{rombach2022high}. We use its encoder $\gE^{\mathrm{AE}}$ to obtain an embedding for image reconstruction $\vx_0^{\mathrm{AE}}$. The second part is the image CLIP~\cite{radford2021learning} (ViT-B/32). It extracts a semantic embedding $\vx_0^{\mathrm{CLIP}}$ of dimension 512. The final latent embedding for images is the concatenation of the outputs from two parts, i.e.,  $\vx_0 = [\vx_0^{\mathrm{AE}}, \vx_0^{\mathrm{CLIP}}]$. Empirically, we found that $\vx_0^{\mathrm{AE}}$ is sufficient for image reconstruction via the image decoder $\gD^{\mathrm{AE}}$ from Stable diffusion and the additional $\vx_0^{\mathrm{CLIP}}$ helps understand the semantics of images in image-to-text generation. We hypothesize that the different roles of the two embeddings are inherently caused by the original objectives, i.e. reconstruction versus semantics alignment with text. 

\textbf{Text encoder-decoder.} As for the text encoder, we employ the same text CLIP as Stable Diffusion~\cite{rombach2022high}. The text CLIP outputs 77 vectors and each is 768-dimensional. To facilitate training, we add an extra linear layer, which reduces the dimension of each vector to 64 to obtain the final text embedding $\vy_0$. We construct the text decoder $\gD^\txt$ based on GPT-2~\cite{radford2019language}. Specifically, GPT-2 takes $\vy_0$ as a prefix embedding~\cite{mokady2021clipcap} and reconstructs the text autoregressively. Freezing the parameters in CLIP, we train the linear layer and fine-tune GPT-2 to reconstruct the input texts, which performs well on reconstruction. We present more training details and the reconstruction results in Appendix~\ref{sec:gpt2}.

\textbf{Remark.} We observe that the latent embeddings of both image and text already have similar and reasonable numerical ranges. Specifically, they are concentrated within the range of $[-2, 2]$ and exhibit approximately normal distributions with comparable mean and variance values (image modality: mean = $0.0269$, standard deviation = $0.7919$; text modality: mean = $0.0127$, standard deviation = $0.5957$). As a result, we did not apply additional normalization to them.
For more modalities, we can similarly convert them to continuous latent features through encoders that have regularization on the latent space. This makes it easy for all modalities to have similar ranges after normalization. 
Besides, obtaining high-quality encoders and decoders is relatively straightforward and can be achieved with a smaller amount of data. For example, the dataset size of the image encoder and decoder is less than 1\% of UniDiffuser's. Therefore, in practice, we can efficiently train high-quality encoders and decoders for each modality at a modest cost if needed.

\subsection{Transformer as Joint Noise Prediction Network}
\label{sec:uvit}

We train a joint noise prediction network on the embeddings obtained in Section~\ref{sec:encode} according to Eq.~{(\ref{eq:j_l2})}.
It is natural to employ a transformer-based backbone in \ours{} to handle inputs from different modalities. In particular, we adopt U-ViT~\cite{bao2022all}, a recently proposed transformer for conditional diffusion models. The original U-ViT is characterized by treating all inputs including the data, the condition, and the timestep as tokens, and employing long skip connections between shallow and deep layers. In \ours{}, we slightly modify U-ViT by treating the two modalities of data and their corresponding timesteps as tokens. Besides, we empirically find that the pre-layer normalization~\cite{xiong2020layer} in the original U-ViT causes overflow easily when trained with mixed precision. A simple solution is to use the post-layer normalization~\cite{vaswani2017attention} and add a layer normalization after concatenating a long skip connection, which stabilizes the training of \ours{}.
We illustrate the backbone in Figure~{\ref{fig:pipeline} (b)} and present more details in Appendix~\ref{sec:detail_uvit}.

\section{Related Work}
\label{sec:related_work}

\textbf{Multi-modal generative modeling.} Many prior work on multi-modal generative modeling can be formalized as learning a conditional distribution. Representative applications include text-to-image generation~\cite{ramesh2021zero,ding2021cogview,ramesh2022hierarchical,nichol2021glide,saharia2022photorealistic,yu2022scaling,gu2022vector,xu2018attngan,rombach2022high}, text-to-video generation~\cite{ho2022imagen}, text-to-speech generation~\cite{chen2020wavegrad,popov2021grad} and image captioning (i.e., image-to-text generation)~\cite{mokady2021clipcap,chen2022analog}. Such models are specially designed for a single task. In addition to learning a conditional distribution, \citet{hu2022unified} aims to learn the joint distribution of image and text data via a discrete diffusion model~\cite{gu2022vector}. However, its scalability is unexplored.

The most related prior work is Versatile Diffusion (VD)~\cite{xu2022versatile}, which employs a multi-flow architecture and is trained for multiple generation tasks in the traditional multi-task framework, which requires multiple feed-forward to compute losses for all tasks and carefully tuned gradient multipliers for different layers during training. In contrast, \ours{} provides an elegant solution based on the insightful  unified view of training diffusion models. As a result, \ours{} is simpler (with a single training loss), more efficient to train (with a single forward-backward per update), and can handle more tasks (able to perform joint sampling) without the need for complex tricks. 
Besides,  \ours{} outperforms VD in both image-to-text and text-to-image generation tasks in terms of the FID and CLIP scores in our experiments (see Section~\ref{sec:exp}), suggesting that the time-condition strategy in \ours{} is statistically more efficient than the multi-task one in VD.

\textbf{Multi-modal representation learning} aims to learn features for different modalities that can be transferred to downstream tasks. Vision-and-language pretraining (VLP) is at the front. VLP can employ different strategies, such as contrastive learning~\cite{radford2021learning}, masked data modeling~\cite{wang2022image}, and a combination of multiple losses~\cite{kim2021vilt,li2022blip,bao2021vlmo}. A transformer is often employed to fuse the two modalities. This work implies that a transformer is also effective for multi-modal generative modeling.

\textbf{Diffusion models} are initially proposed by \citet{sohl2015deep}. Recently, \citet{ho2020denoising} introduce a noise prediction formulation, and \citet{song2020score} introduce a stochastic differential equation formulation for learning diffusion models. Diffusion models are able to generate high-quality images~\cite{dhariwal2021diffusion},audios~\cite{chen2020wavegrad,kong2020diffwave}, videos~\cite{ho2022video}, point clouds~\cite{luo2021diffusion} and molecular conformations~\cite{hoogeboom2022equivariant,bao2022equivariant}. Other improvements in diffusion models include fast sampling~\cite{song2020denoising,bao2022analytic,salimans2022progressive,lu2022dpm,lu2022dpmpp} and improved training and sampling techniques~\cite{nichol2021improved,song2021maximum,kingma2021variational,vahdat2021score,zhao2022egsde,bao2022estimating,lu2022maximum,karras2022elucidating}.

\section{Experiments}
\label{sec:exp}

We present the experimental setup in Section~\ref{sec:setup}. We show the ability of \ours{} to perform multiple generation tasks and directly compare it with existing large models in Section~\ref{sec:main_results}. We further demonstrate that \ours{} naturally supports applications like data variation, blocked Gibbs sampling between modalities (see Section~\ref{sec:variation}), and interpolation between images in the wild (see Section~\ref{sec:interpolation}).

\begin{figure}[t]
    \centering
    \includegraphics[width=0.9\columnwidth]{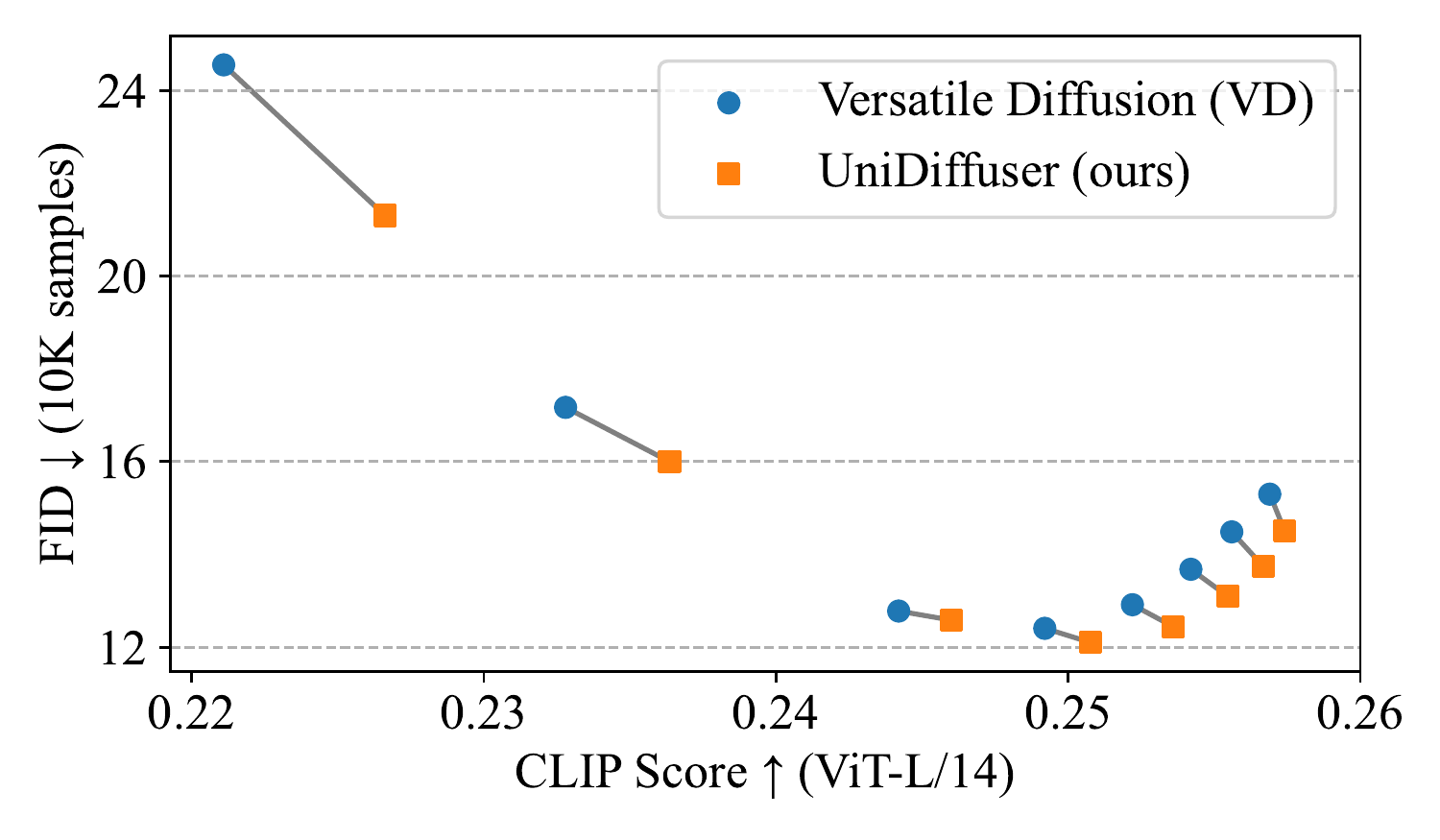}
    \caption{\textbf{Comparing \ours{} and VD in text-to-image generation.} We connect the results with the same scale in CFG. \ours{} consistently outperforms VD in all settings w.r.t. both the CLIP score $\uparrow$ (horizontal axis) and FID $\downarrow$  (vertical axis).}
    \label{fig:t2i_fid_clip}
\end{figure}

\subsection{Setup}
\label{sec:setup}

\textbf{Dataset.} We use three subsets of LAION-5B~\cite{schuhmann2022laion} following Stable Diffusion~\cite{rombach2022high}. The first one is laion2B-en, which contains around 2B image-text pairs with English captions. The second one is laion-high-resolution, which contains around 170M image-text pairs with image resolution $\geq$1024 and multilingual captions. The third one is laion-aesthetics v2 5+, which is a subset of laion2b-en containing around 600M image-text pairs with high visual quality. Following Stable Diffusion, we additionally filter laion-aesthetics v2 5+ to images with resolution $\geq$512 and an estimated watermark probability $<$0.5, leading to around 193M preserved pairs. For image normalization, we follow the standard practice in diffusion models by normalizing the image values from the range of $[0, 255]$ to $[-1, 1]$. Since the texts in LAION-5B are quite noisy, we further clean texts in the laion-aesthetics v2 5+ subset by removing URLs, HTML tags, emails, contents in brackets, quotes except \texttt{'s}, and symbols except \texttt{, . ? !}. Before inputting the text into CLIP, we tokenize the preprocessed text using CLIP's built-in tokenizer, which is based on byte-level Byte-Pair-Encoding~\cite{radford2021learning}.

\textbf{Training and Sampling.} The training is multiple-staged following Stable Diffusion~\cite{rombach2022high}. In the first stage, we train 250K steps at 256$\times$256 resolution on laion2B-en with a batch size of 11264 and 5K warm-up steps. 
In the second stage, we fine-tune the model with 200K steps at 512$\times$512 resolution on laion-high-resolution with a batch size of 2112 and 5K warm-up steps.
In the last stage, we resume from the last checkpoint of the second stage (including both weights of the model and states of the optimizer), and train 220K steps at 512$\times$512 resolution on laion-aesthetics v2 5+ with a batch size of 2112.
Following \citet{bao2022all}, we use the AdamW optimizer~\cite{loshchilov2017decoupled} with a learning rate of 2e-4, a weight decay of 0.03 and running coefficients of $(\beta_1, \beta_2) = (0.9, 0.9)$ in all stages.
We reduce the learning rate by a factor of 10 and continue training whenever the validation loss does not decrease. We train with mixed precision for efficiency.
When U-ViT is trained at 256$\times$256 resolution, we interpolate the positional embeddings related to images via bilinear interpolation. The training takes around 28 days on 88 A100 (80GB) GPUs. We use DPM-Solver~\cite{lu2022dpm,lu2022dpmpp} with 50 steps in all experiments.

\textbf{Baseline.} To our knowledge, Versatile Diffusion (VD)~\cite{xu2022versatile} is the most direct competitor for general-purpose multi-modal generation (see details in Section~\ref{sec:related_work}). We directly compare to VD in all experiments if possible. The results of VD are reproduced by us upon official code because there is no quantitative result in the original paper.

\textbf{Evaluation.} For text-to-image generation, we report the FID~\cite{heusel2017gans} and CLIP score~\cite{radford2021learning} on the MS-COCO validation set~\cite{lin2014microsoft} to measure the image fidelity and image-text alignment respectively. Following the literature, we randomly draw 10K and 30K prompts from the MS-COCO validation set to calculate FID and the CLIP score on generated images.
For image-to-text generation, we report the CLIP score to measure the image-text alignment. Specifically, we randomly draw 10K images to calculate the score on generated texts.

\begin{figure}[t]
    \centering
    \includegraphics[width=0.9\columnwidth]{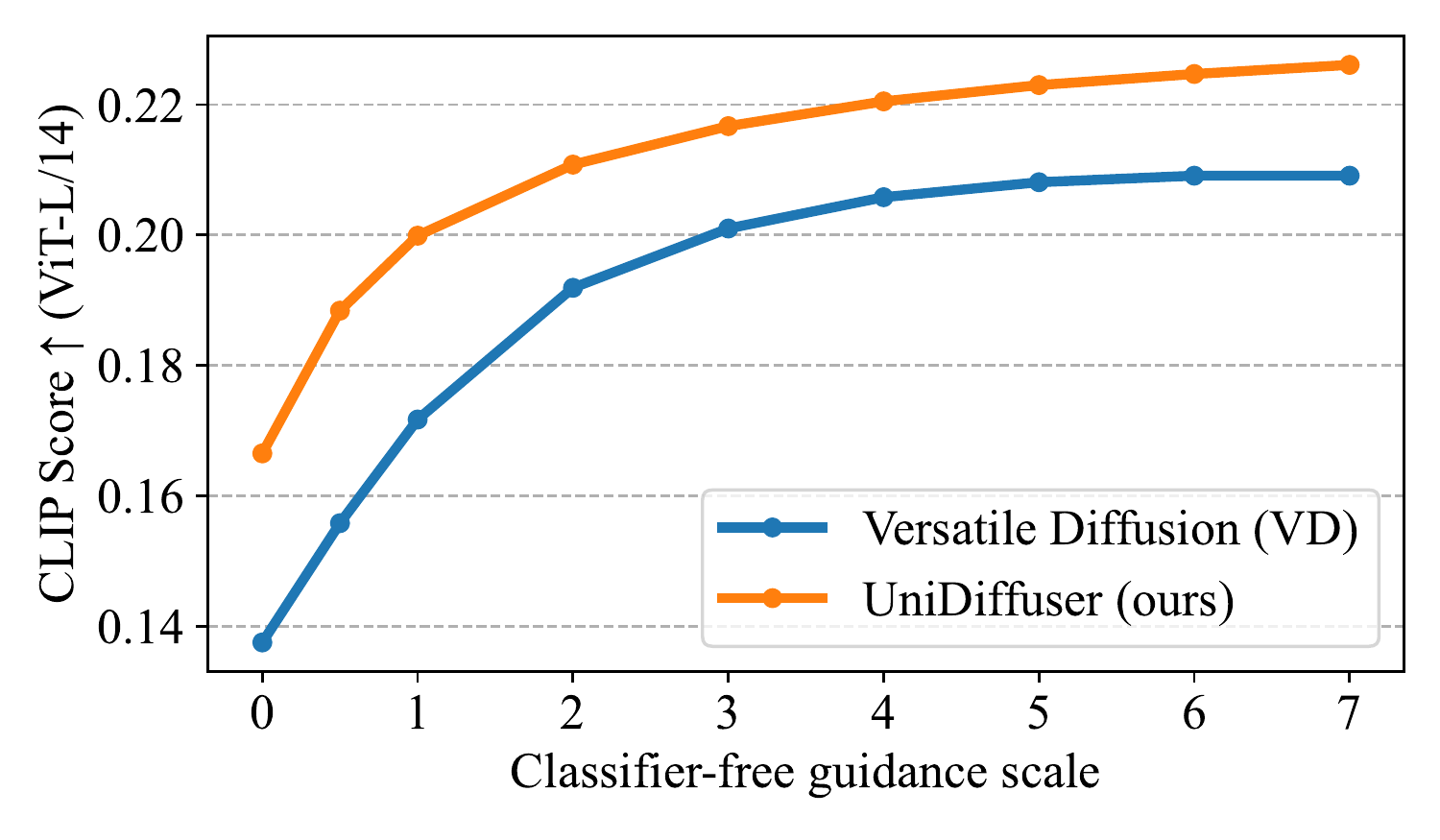}
    \caption{\textbf{Comparing \ours{} and VD in image-to-text generation.} \ours{} consistently outperforms VD with the same CFG scale (horizontal axis)  w.r.t. the CLIP score $\uparrow$ (vertical axis).}
    \label{fig:i2t_clip}
\end{figure}

\begin{figure}[t]
    \centering
   \includegraphics[width=\columnwidth]{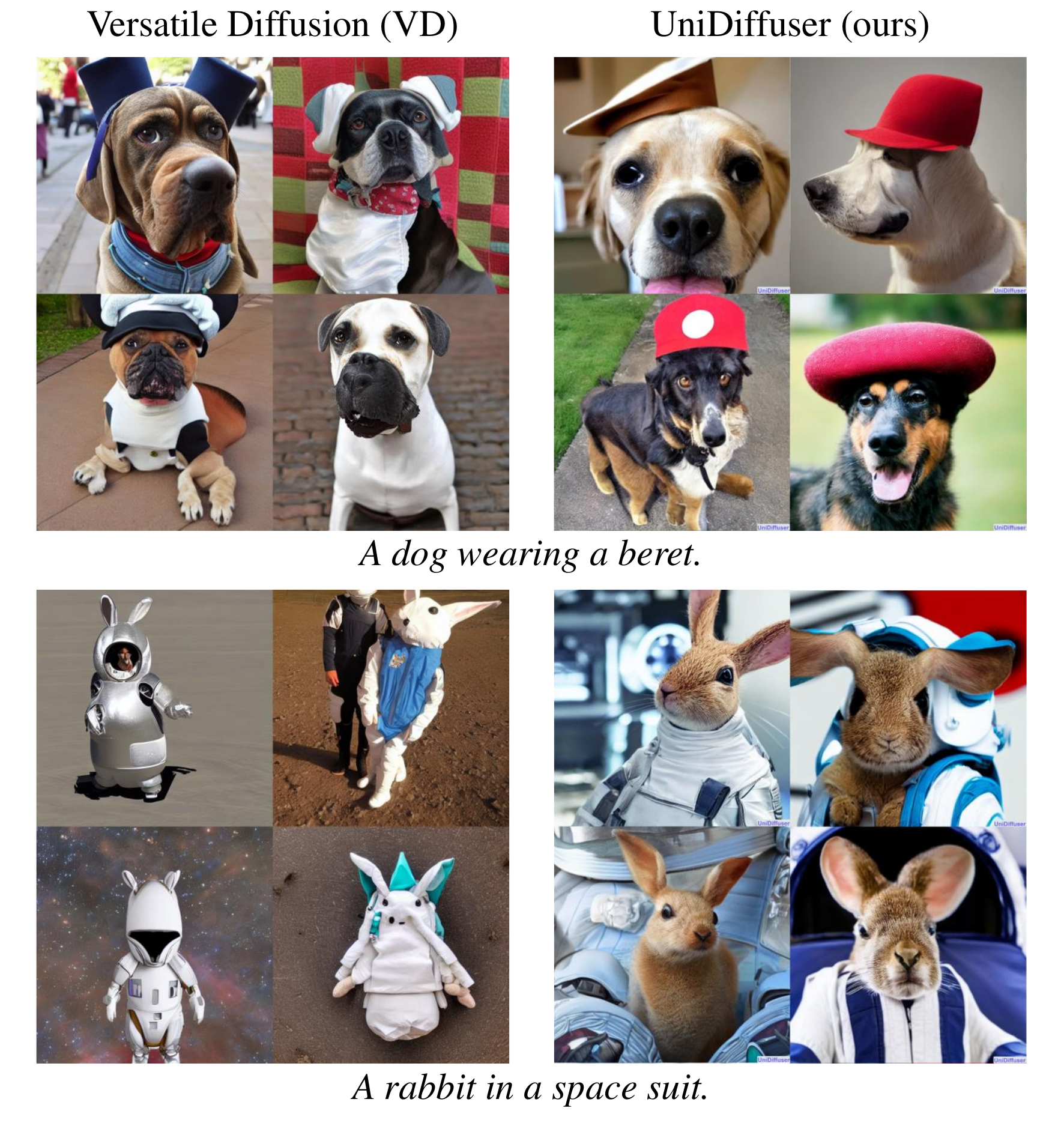}
    \caption{\textbf{Random samples of \ours{} and VD on text-to-image generation.} \ours{} produces semantically correct images given representative prompts while VD does not.}
    \label{fig:compare_t2i}
\end{figure}

\begin{table}[t]
    \caption{\textbf{Zero-shot FID $\downarrow$ on the MS-COCO validation set}. $^\dagger$ marks results produced by us upon official implementation and other results are taken from the corresponding references. We report the results of \ours{} and VD with a scale of 3 in CFG, which is the best choice for both models according to Figure~\ref{fig:t2i_fid_clip}.}\label{tab:fid}
    \vskip 0.1in
    \centering
    \begin{tabular}{lr}
    \toprule
        Model &  FID $\downarrow$ \\
    \midrule
    \emph{Bespoken models} \\
    ~~~~GLIDE~\cite{nichol2021glide} & 12.24 \\
        ~~~~Make-A-Scene~\cite{gafni2022make} & 11.84 \\
        ~~~~DALL$\cdot$E 2~\cite{ramesh2022hierarchical} & 10.39 \\
        ~~~~Stable Diffusion$^\dagger$~\cite{rombach2022high} & 8.59 \\
        ~~~~Imagen~\cite{saharia2022photorealistic} & 7.27 \\
        ~~~~Parti~\cite{yu2022scaling} & \textbf{7.23} \\
    \midrule
    \emph{General-purpose models} \\
        ~~~~Versatile Diffusion$^\dagger$~\cite{xu2022versatile} & 10.09 \\
        ~~~~\ours{} (\textbf{ours}) & \textbf{9.71} \\
    \bottomrule
    \end{tabular}
\end{table}

\subsection{Main Results}
\label{sec:main_results}

We first systematically compare with the most direct baseline Versatile Diffusion (VD), which is a general-purpose generative model, in both text-to-image and image-to-text generation. Quantitatively, \ours{} outperforms VD consistently in both tasks under all metrics and guidance CFG scales, as presented in Figure~\ref{fig:t2i_fid_clip} and Figure~\ref{fig:i2t_clip}. The empirical results demonstrate the effectiveness (in addition to the simplicity, efficiency, and generality) of \ours{} compared to VD (see details in Section~\ref{sec:related_work}). Qualitatively, Figure~\ref{fig:compare_t2i} presents samples in text-to-image generation, and \ours{} aligns image and text better than VD. See more results including image-to-text generation
in Appendix~\ref{sec:cmp_example}.

% We firstly compare the text-to-image generation performance between \ours{} and VD. In Figure~\ref{fig:t2i_fid_clip}, we plot the FID-CLIP score trade-off curve by sweeping the classifier-free guidance scale in \{0.5, 1.0, 2.0, 3.0, 4.0, 5.0, 6.0, 7.0\}. \ours{} outperforms VD at all scales.
% Then we compare the image-to-text performance between \ours{} and VD. In Figure~\ref{fig:i2t_clip}, we plot the CLIP score at different guidance scales. \ours{} still outperforms VD at all scales.

We also compare with bespoken systems designed for text-to-image generation w.r.t. zero-shot FID on MS-COCO in Table~\ref{tab:fid}. Although \ours{} is designed to handle multiple generation tasks, its performance on the single text-to-image generation task is comparable to bespoken diffusion models such as Stable Diffusion and outperforms famous diffusion models like DALL$\cdot$E 2.

Finally, we present examples of joint, conditional, and unconditional generation in Figure~\ref{fig:demos} (a-e) to show the generality of \ours{}. See more examples in Appendix~\ref{sec:more}.

\subsection{Data Variation and Gibbs Sampling}
\label{sec:variation}
\ours{} naturally supports applications such as image variation and text variation. For example, given a source image, we can firstly perform image-to-text generation to obtain a description of the image, and then perform text-to-image generation with this description as input to obtain a new image with similar semantics but different contents.
In Figure~\ref{fig:demos} (f-g), we present examples on image and text variation. 
Furthermore, we can perform blocked Gibbs sampling to see how images and texts are translated to each other by chaining conditional distributions modeled by \ours{}.
We present examples in Figure~\ref{fig:demos} (h). More samples on data variation and blocked Gibbs can be found in Appendix~\ref{sec:more}.

\subsection{Interpolation between Two Images in the Wild}
\label{sec:interpolation}

\ours{} can also perform interpolation between two images in the wild. Specifically, we firstly perform image-to-text generation to obtain the latent text embeddings of the two images via the deterministic DPM-Solver with the same Gaussian noise as the initial state for both images. Then we perform a noise injection process via DPM-Solver to get a noisy version of the latent image embeddings given the two latent text embeddings. We perform spherical linear interpolation between the latent text embeddings and the noisy version of the latent image embeddings to obtain intermediate states. Finally, with the text intermediate states as the condition and the image intermediate states as the initial state, we generate the final images by DPM-solver. See Appendix~\ref{sec:interpolation_alg} for a formalized algorithm of the interpolation procedure. We present examples in Figure~{\ref{fig:demos} (i)} and more examples can be found in Appendix~\ref{sec:more}.

\section{Conclusion}

We propose \ours{}, a general-purpose multi-modal probabilistic framework based on insights of unifying training of diffusion models for different distributions. \ours{} is able to perform various generation tasks via one model with minimal modification of the original diffusion models. Empirical results on image-text data show the effectiveness of \ours{} compared to large existing models. \ours{} also enables semi-supervised learning and learning on more modalities, which are left as future work. Currently, the text generated by our implementation is not that smooth, mainly because the text data is noisy.

UniDiffuser has high potential to improve multiple tasks: by fitting multiple tasks with one single transformer network, UniDiffuser can be much easier to further improve all tasks simultaneously (e.g., by increasing parameter scale and data scale) and maintain under the large-scale pre-training regime. Any further improvement/optimization of the underlying single network can seamlessly benefit all tasks.

\textbf{Social Impact:} We believe \ours{} can advance real-world applications with generated content due to its generality. However, it is worth noting that large-scale multi-modal generative models may have consequences like ``deep-fakes''. We watermark all images sampled from the model and will provide a systematical protocol to relieve the problem before releasing the code and model.

% Authors are kindly asked to make their submissions as accessible as possible for everyone including people with disabilities and sensory or neurological differences.
% Tips of how to achieve this and what to pay attention to will be provided on the conference website \url{http://icml.cc/}.

% \section*{Software and Data}

% If a paper is accepted, we strongly encourage the publication of software and data with the
% camera-ready version of the paper whenever appropriate. This can be
% done by including a URL in the camera-ready copy. However, \textbf{do not}
% include URLs that reveal your institution or identity in your
% submission for review. Instead, provide an anonymous URL or upload
% the material as ``Supplementary Material'' into the CMT reviewing
% system. Note that reviewers are not required to look at this material
% when writing their review.

\section*{Acknowledgements}

This work was supported by NSF of China Projects (Nos. 62061136001, 61620106010, 62076145, U19B2034, U1811461, U19A2081, 6197222);  Beijing Outstanding Young Scientist Program NO. BJJWZYJH012019100020098; a grant from Tsinghua Institute for Guo Qiang; the High Performance Computing Center, Tsinghua University; the Fundamental Research Funds for the Central Universities, and the Research Funds of Renmin University of China (22XNKJ13). C. Li was also sponsored by Beijing Nova Program. J.Z was also supported by the New Cornerstone Science Foundation through the XPLORER PRIZE.


\bibliography{example_paper}
\bibliographystyle{icml2023}

%%%%%%%%%%%%%%%%%%%%%%%%%%%%%%%%%%%%%%%%%%%%%%%%%%%%%%%%%%%%%%%%%%%%%%%%%%%%%%%
%%%%%%%%%%%%%%%%%%%%%%%%%%%%%%%%%%%%%%%%%%%%%%%%%%%%%%%%%%%%%%%%%%%%%%%%%%%%%%%
% APPENDIX
%%%%%%%%%%%%%%%%%%%%%%%%%%%%%%%%%%%%%%%%%%%%%%%%%%%%%%%%%%%%%%%%%%%%%%%%%%%%%%%
%%%%%%%%%%%%%%%%%%%%%%%%%%%%%%%%%%%%%%%%%%%%%%%%%%%%%%%%%%%%%%%%%%%%%%%%%%%%%%%
\newpage
\appendix
\onecolumn

\section{More Examples}
\label{sec:more}

\begin{figure}[H]
    \centering
    \begin{tabular}{ccc}
         \includegraphics[width=0.3\textwidth]{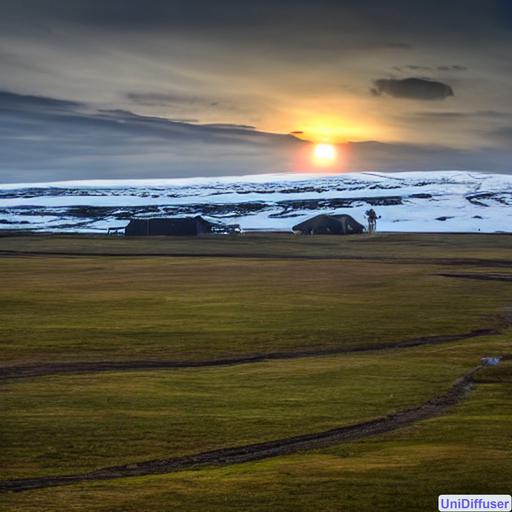} & \includegraphics[width=0.3\textwidth]{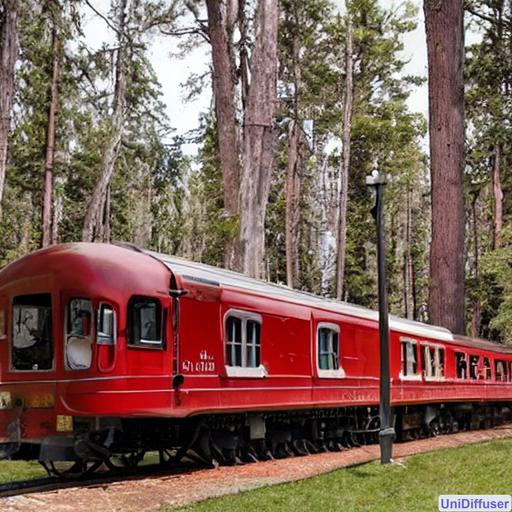} & \includegraphics[width=0.3\textwidth]{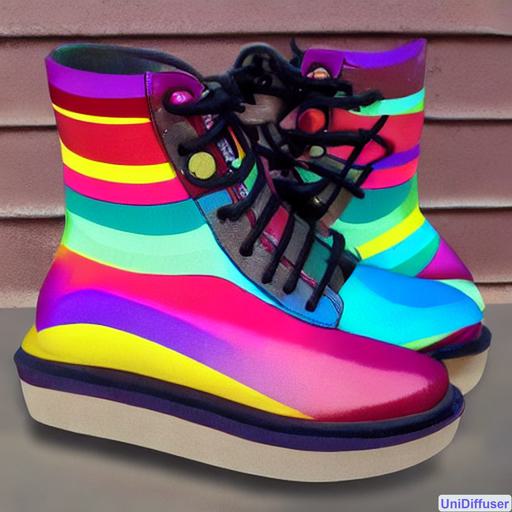} \\
        \makecell{A dramatic sunset over rural fields \\ with Icelandic hills Iceland} & \makecell{An old red electric rail train \\ in Durango, Colorado} & Colourful Rainbow Triangle Shoes \vspace{.2cm} \\
        \includegraphics[width=0.3\textwidth]{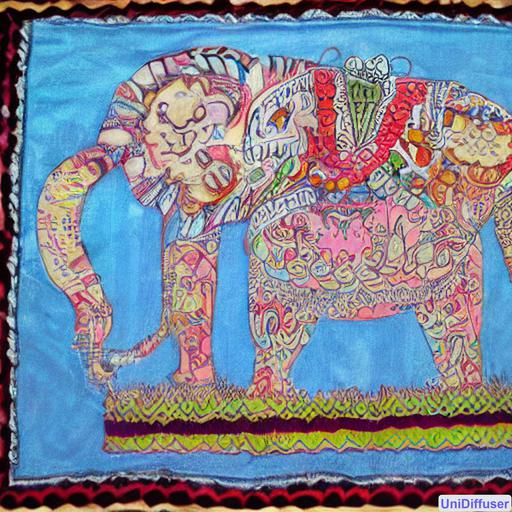} & \includegraphics[width=0.3\textwidth]{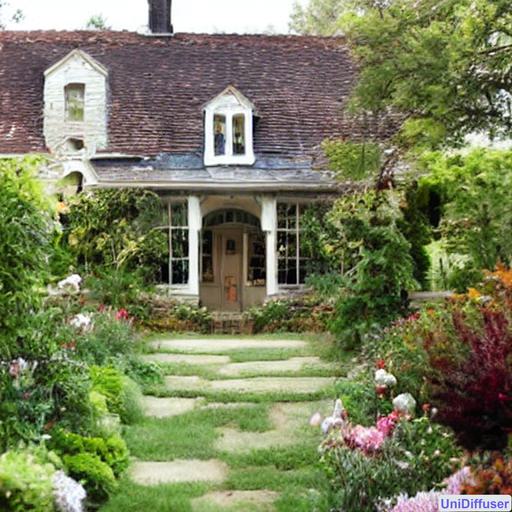} & \includegraphics[width=0.3\textwidth]{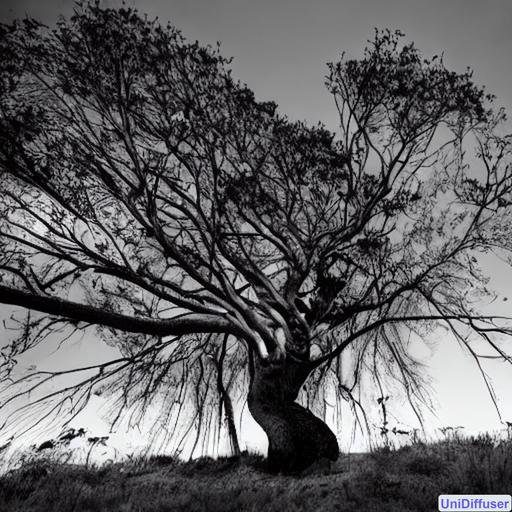} \\
        Elephant Tapestry & English Country Garden Design & \makecell{black and white photography \\ of big tree} \vspace{.2cm} \\
        \includegraphics[width=0.3\textwidth]{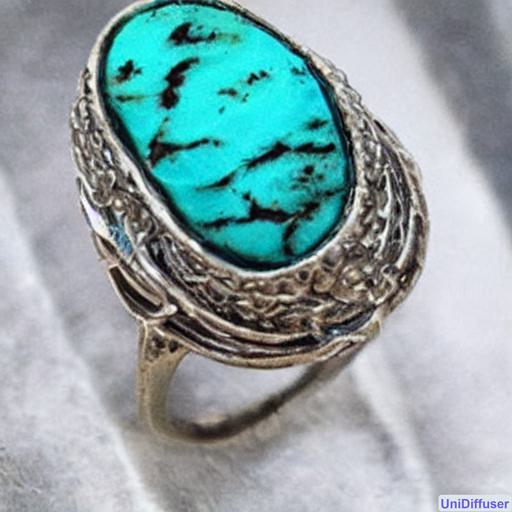} & \includegraphics[width=0.3\textwidth]{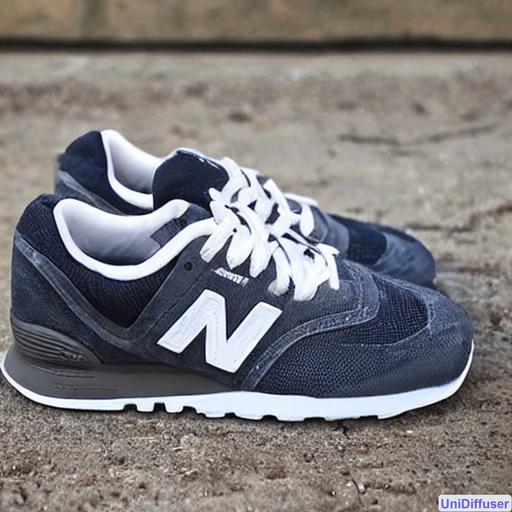} & \includegraphics[width=0.3\textwidth]{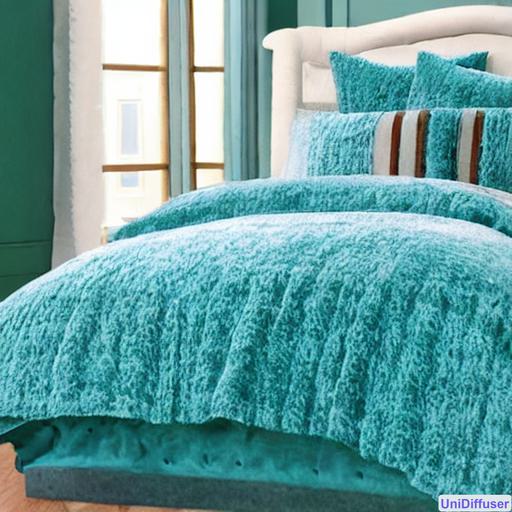} \\
        \makecell{Turquoise Vintage Style \\ Handmade Statement Ring} & new balance 997 sneakers & Teal Chenille Bedspread Sets \\
    \end{tabular}
    \caption{Selected samples of \ours{} on joint generation of image and text.}
    \label{fig:joint_examples}
\end{figure}

\begin{figure}[H]
    \centering
    \begin{tabular}{ccc}
         \includegraphics[width=0.3\textwidth]{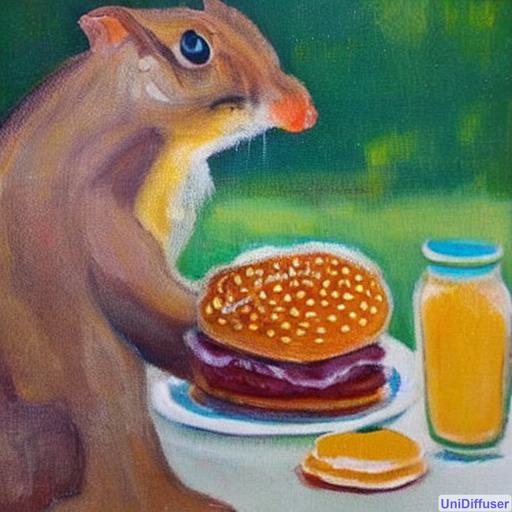} & \includegraphics[width=0.3\textwidth]{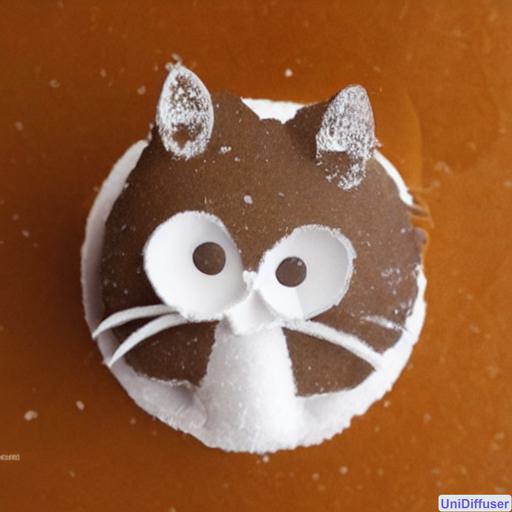} & \includegraphics[width=0.3\textwidth]{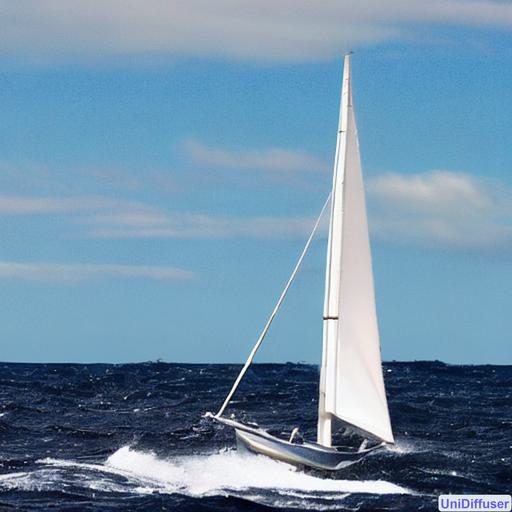} \\
        \makecell{A painting of a squirrel \\ eating a burger} & A cute cat made of sugar & \makecell{A sailboat is sailing \\ the Atlantic Ocean} \vspace{.2cm} \\
        \includegraphics[width=0.3\textwidth]{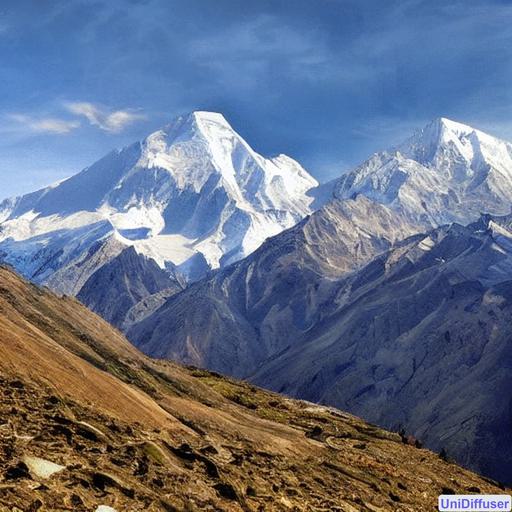} & \includegraphics[width=0.3\textwidth]{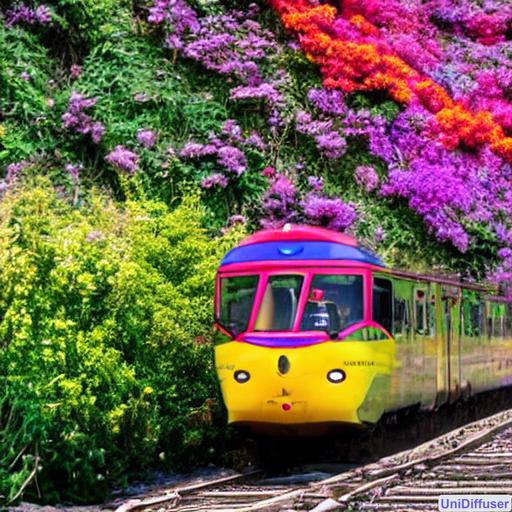} & \includegraphics[width=0.3\textwidth]{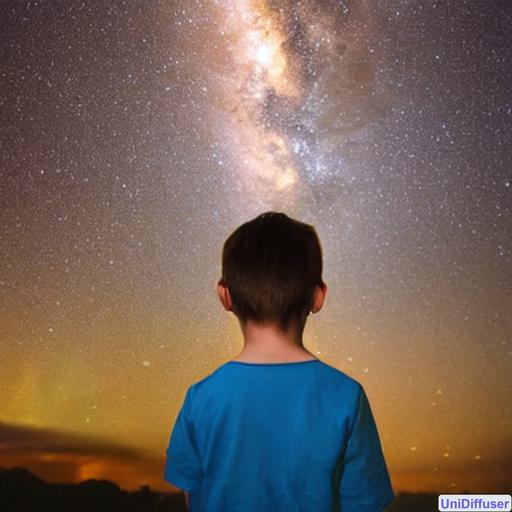} \\
        Beautiful view of the Himalayas & \makecell{A colorful train \\ passes through the flowers} & A boy is looking at the Milky Way \vspace{.2cm} \\
        \includegraphics[width=0.3\textwidth]{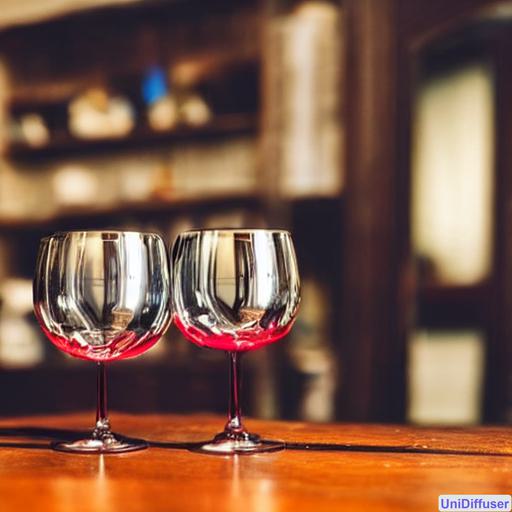} & \includegraphics[width=0.3\textwidth]{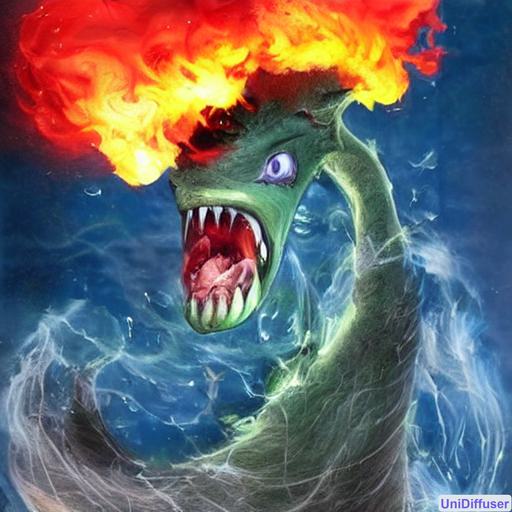} & \includegraphics[width=0.3\textwidth]{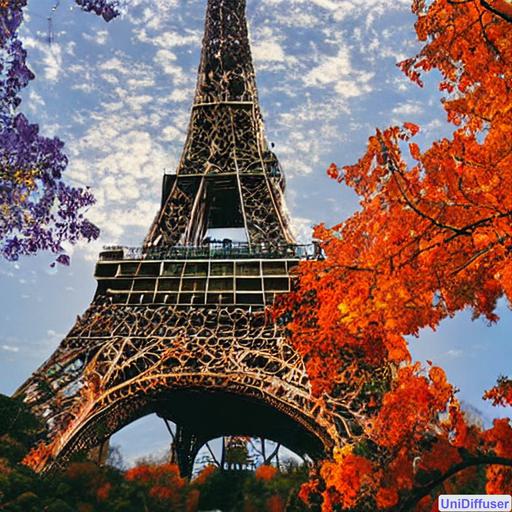} \\
        \makecell{A couple of glasses \\ are sitting on a table} & A fire-breathing monster & \makecell{The beautiful scenery of \\ the Eiffel Tower in Paris} \\
    \end{tabular}
    \caption{Selected samples of \ours{} on text-to-image generation.}
    \label{fig:t2i_examples}
\end{figure}

\begin{figure}[H]
    \centering
    \begin{tabular}{ccc}
         \includegraphics[width=0.3\textwidth]{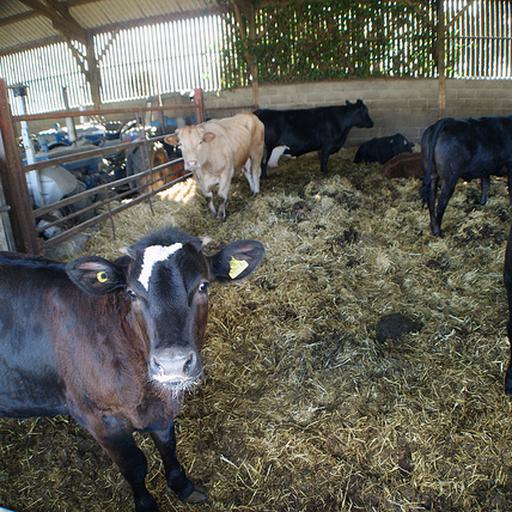} & \includegraphics[width=0.3\textwidth]{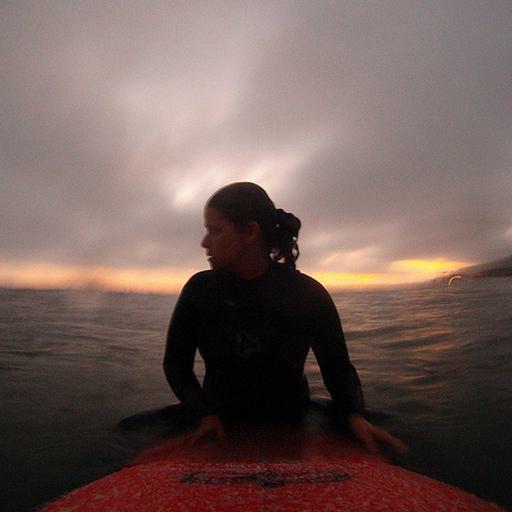} & \includegraphics[width=0.3\textwidth]{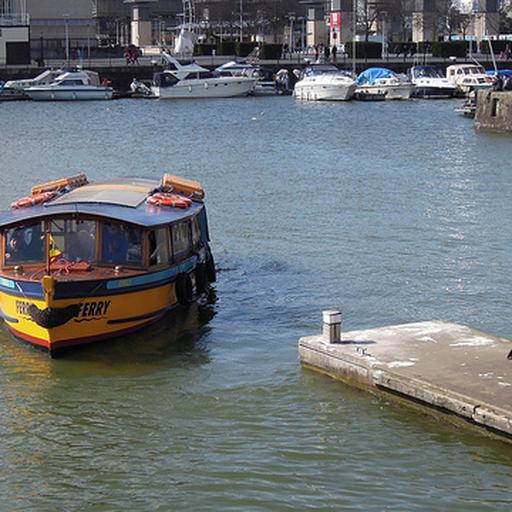} \\
        \makecell{\textit{A herd of cows standing}\\ \textit{inside a shed}} & \makecell{\textit{A woman sits on a kayak of } \\ \textit{Patagonia Argentina sea} \\ \textit{with a sunset clouds} \\ \textit{in the air after the rain.}} & \makecell{\textit{Blue and orange boat} \\ \textit{at the port, Swansea.}} \vspace{.3cm} \\
        \includegraphics[width=0.3\textwidth]{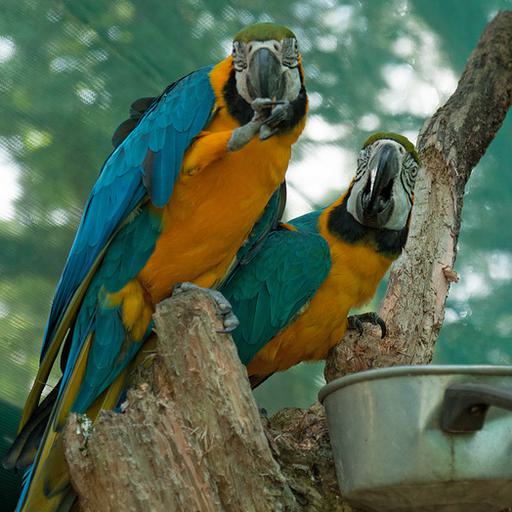} & \includegraphics[width=0.3\textwidth]{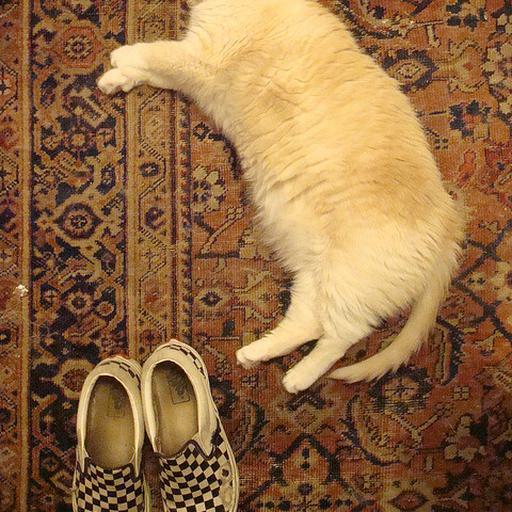} & \includegraphics[width=0.3\textwidth]{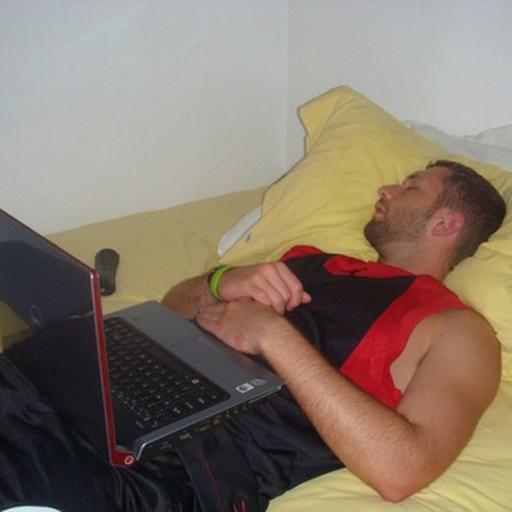} \\
        \makecell{\textit{Blue and yellow macaws sitting} \\ \textit{in a tree precariously}} & \textit{Cat resting on the rug} & \makecell{\textit{Man sleeping on bed} \\ \textit{with his laptop}} \vspace{.3cm} \\
        \includegraphics[width=0.3\textwidth]{imgs/i2t/Red_London_Bus.jpg} & \includegraphics[width=0.3\textwidth]{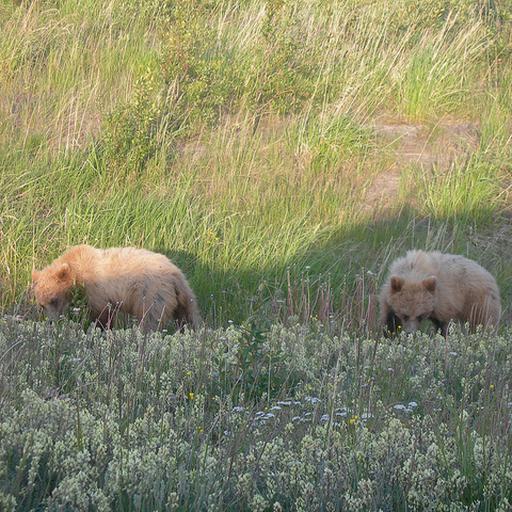} & \includegraphics[width=0.3\textwidth]{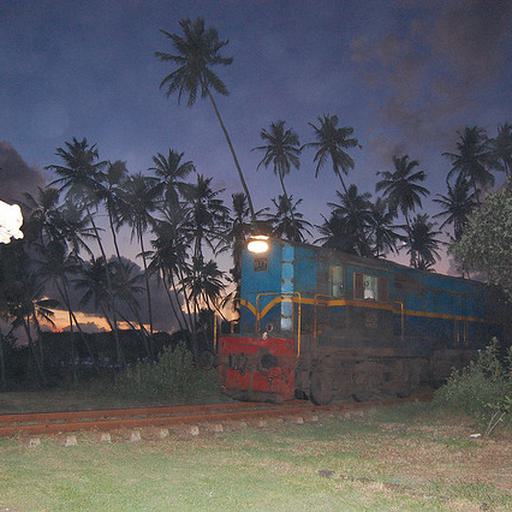} \\
        \textit{Red London Bus} & \textit{Two Grizzly} & \makecell{\textit{Train tracks with} \\ \textit{palm trees at night}} \\
    \end{tabular}
    \caption{Selected samples of \ours{} on image-to-text generation.}
    \label{fig:i2t_examples}
\end{figure}

\begin{figure}[H]
    \centering
    \begin{tabular}{ccc}
         \includegraphics[width=0.3\textwidth]{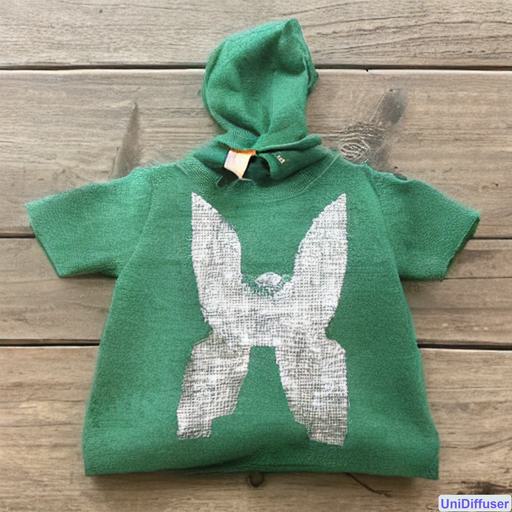} & \includegraphics[width=0.3\textwidth]{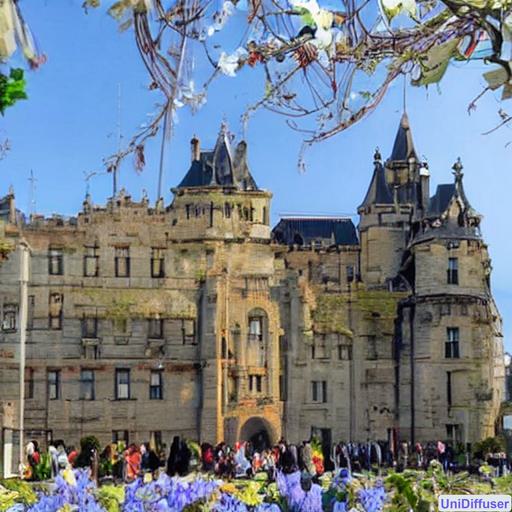} & \includegraphics[width=0.3\textwidth]{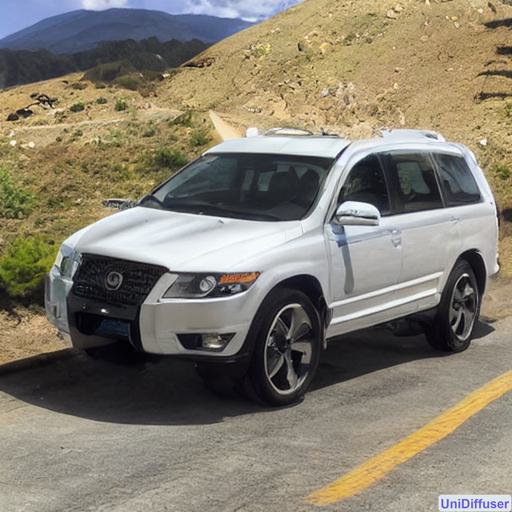} \\
        \includegraphics[width=0.3\textwidth]{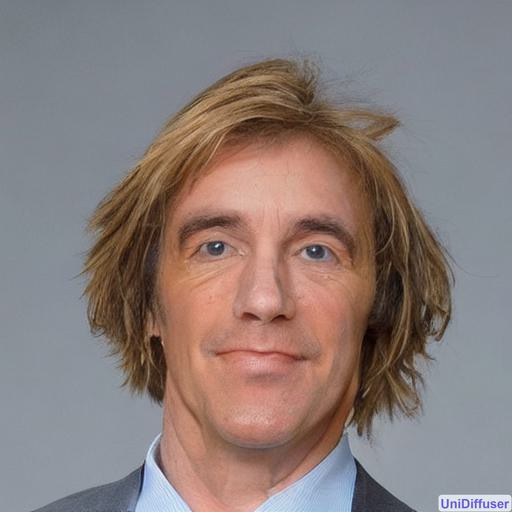} & \includegraphics[width=0.3\textwidth]{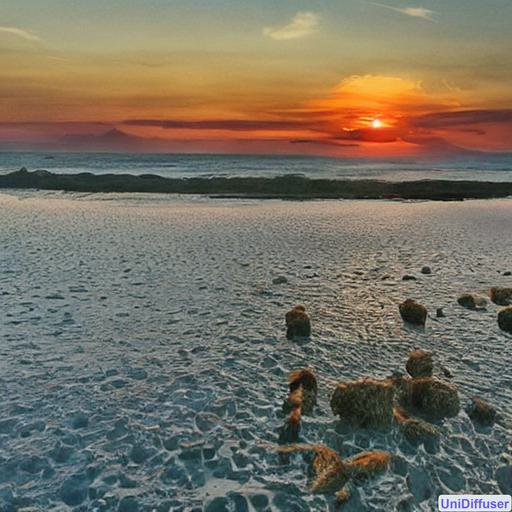} & \includegraphics[width=0.3\textwidth]{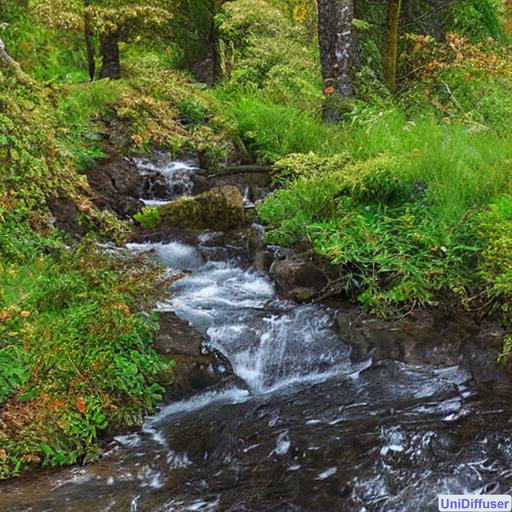} \\
        \includegraphics[width=0.3\textwidth]{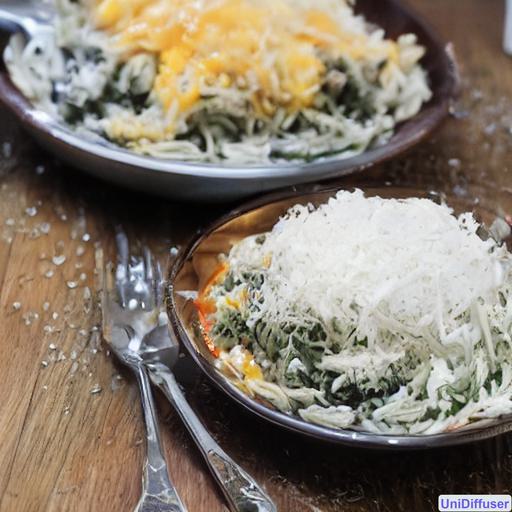} & \includegraphics[width=0.3\textwidth]{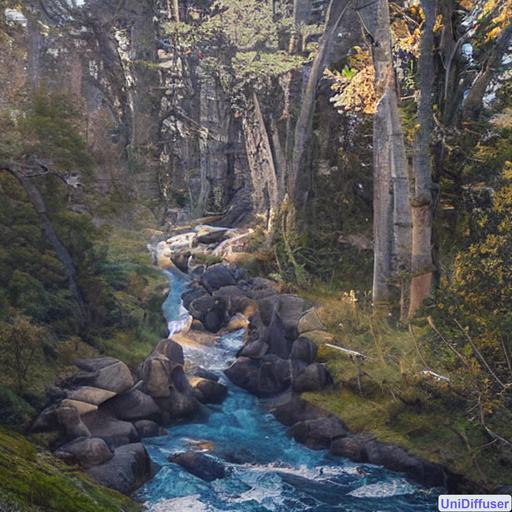} & \includegraphics[width=0.3\textwidth]{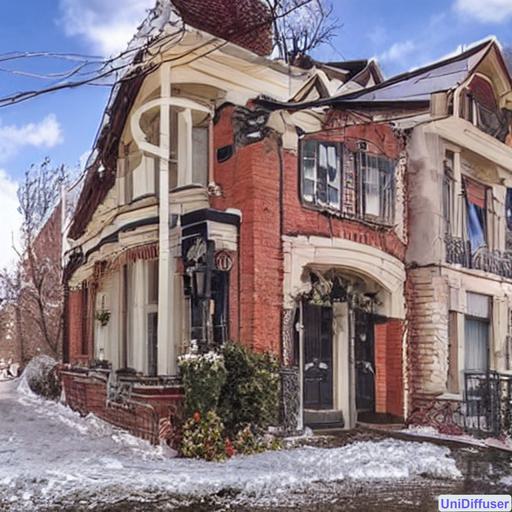} \\
    \end{tabular}
    \caption{Selected samples of \ours{} on unconditional image generation.}
    \label{fig:i_examples}
\end{figure}

\begin{figure}[H]
    \centering
    \begin{tabular}{cc|cc}
        Input & Output & Input & Output \\
         \includegraphics[width=0.22\textwidth]{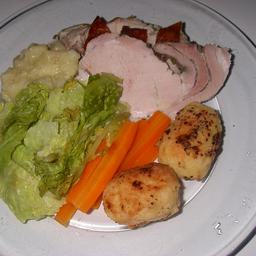} & \includegraphics[width=0.22\textwidth]{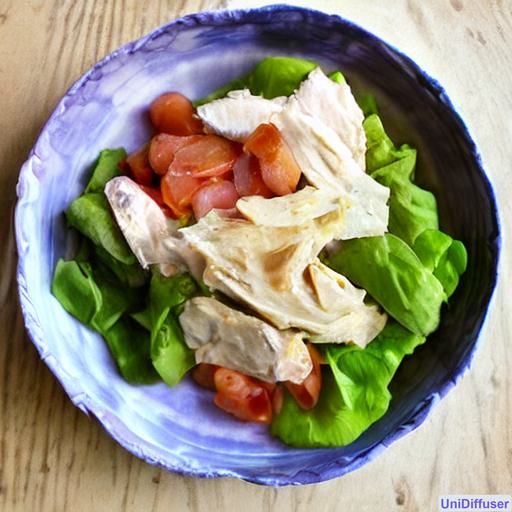} & \includegraphics[width=0.22\textwidth]{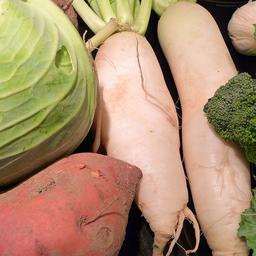} & \includegraphics[width=0.22\textwidth]{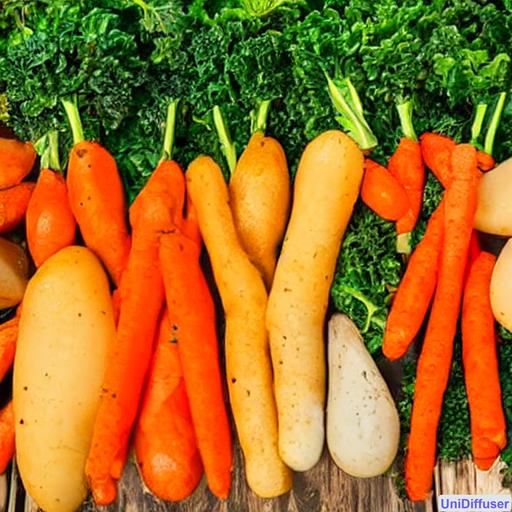}\\
        \includegraphics[width=0.22\textwidth]{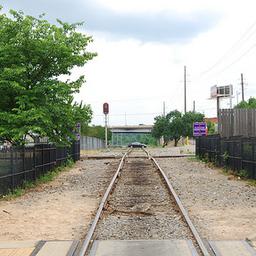} & \includegraphics[width=0.22\textwidth]{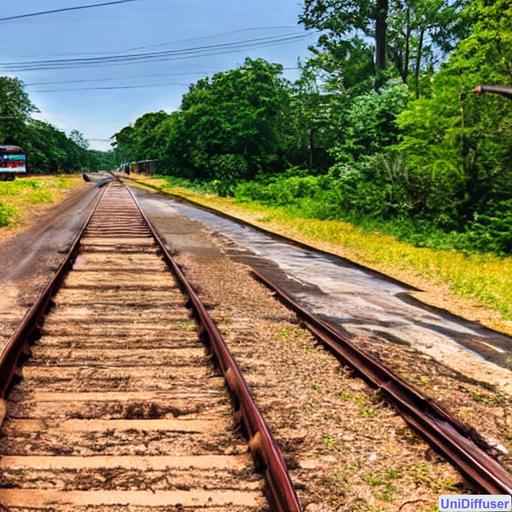} & \includegraphics[width=0.22\textwidth]{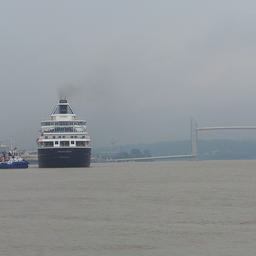} & \includegraphics[width=0.22\textwidth]{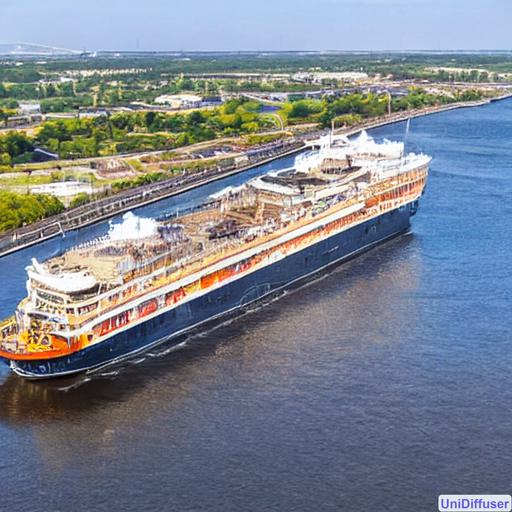}\\
        \includegraphics[width=0.22\textwidth]{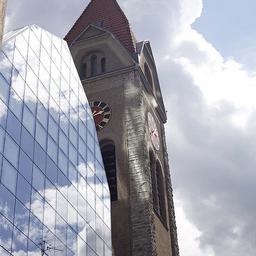} & \includegraphics[width=0.22\textwidth]{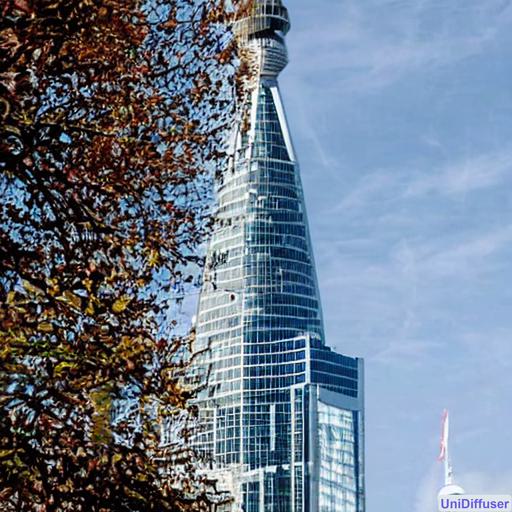} & \includegraphics[width=0.22\textwidth]{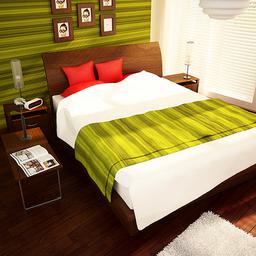} & \includegraphics[width=0.22\textwidth]{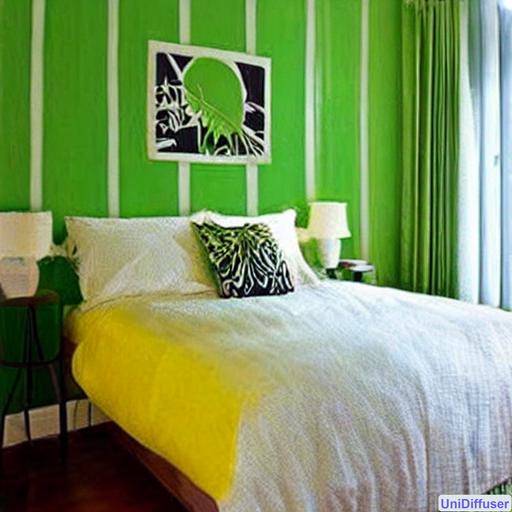}\\
        \includegraphics[width=0.22\textwidth]{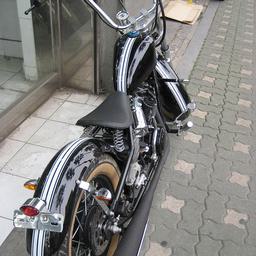} & \includegraphics[width=0.22\textwidth]{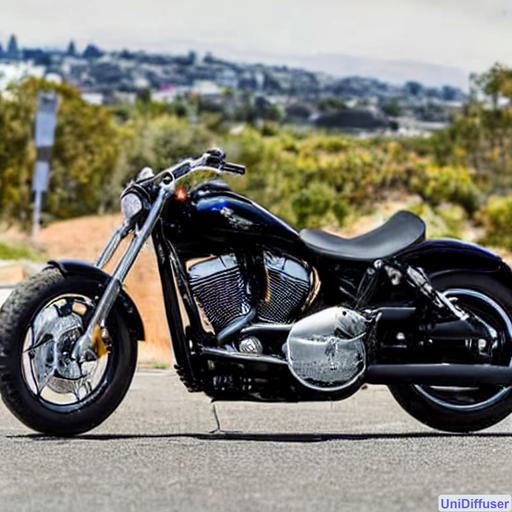} & \includegraphics[width=0.22\textwidth]{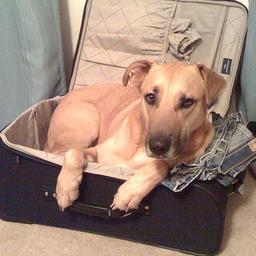} & \includegraphics[width=0.22\textwidth]{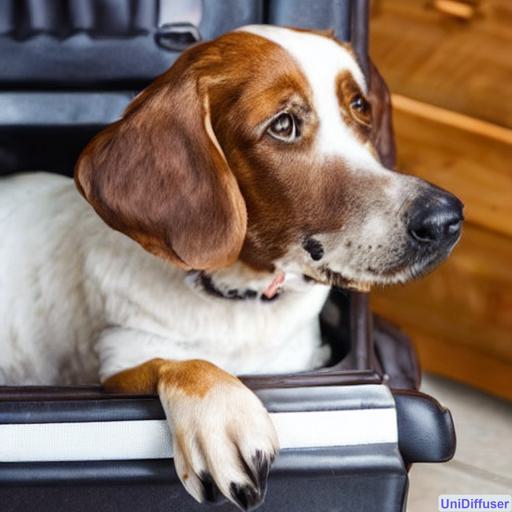}\\
    \end{tabular}
    \caption{Selected samples of \ours{} on image variation.}
    \label{fig:i2t2i_examples}
\end{figure}

\begin{figure}[H]
    \centering
    \includegraphics[width=\linewidth]{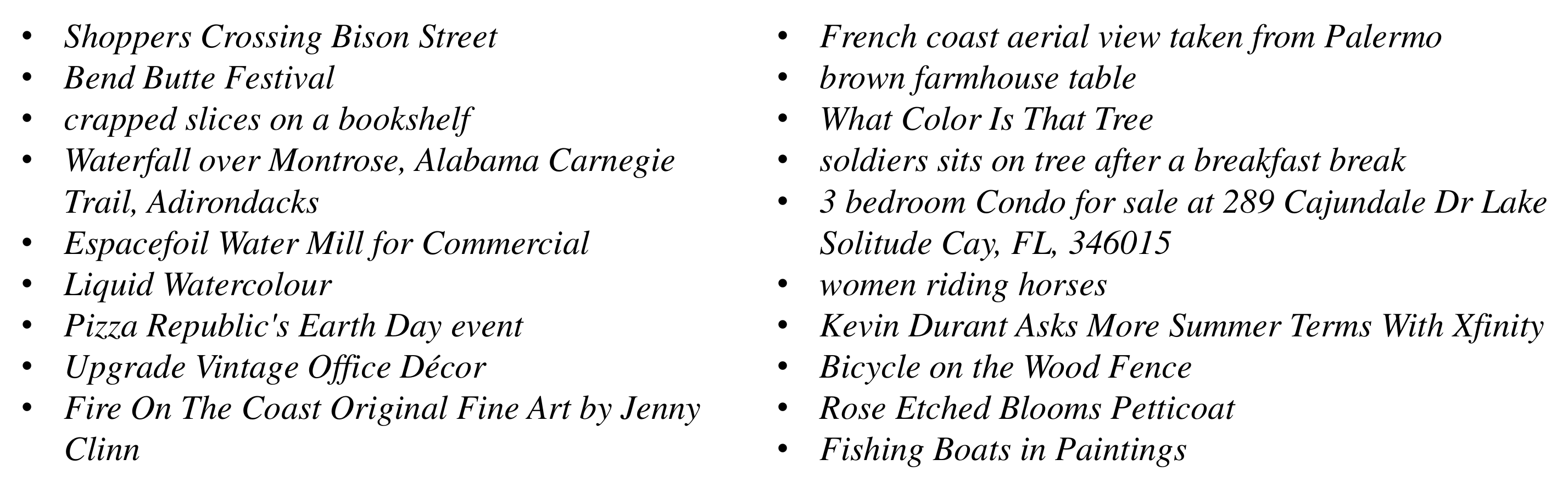}
    \caption{Selected samples of \ours{} on unconditional text generation.}
    \label{fig:t_examples}
\end{figure}

\begin{figure}[H]
    \centering
    \includegraphics[width=\linewidth]{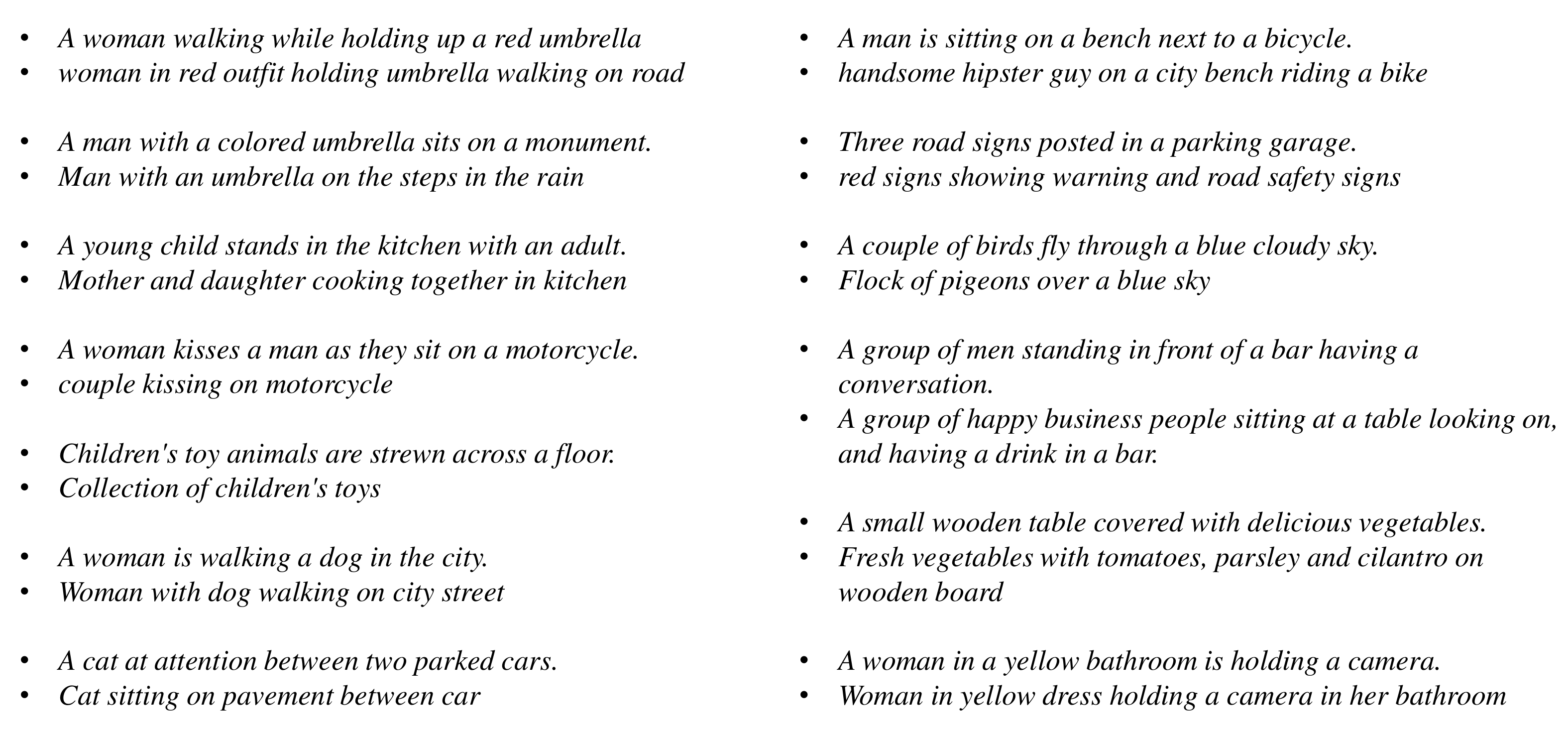}
    \caption{Selected samples of \ours{} on text variation.}
    \label{fig:t2i2t_examples}
\end{figure}

\begin{figure}[H]
    \centering
    \includegraphics[width=\linewidth]{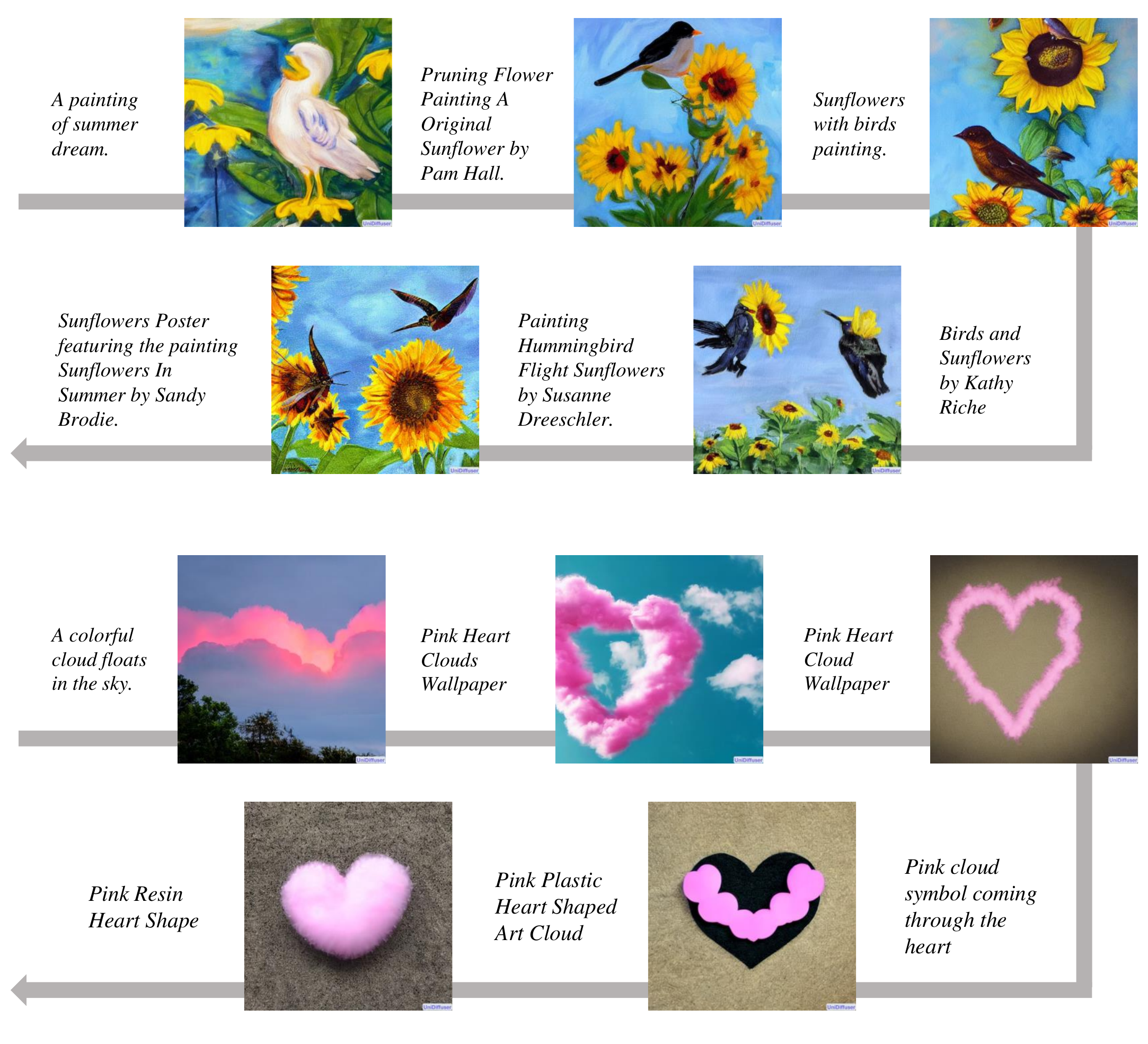}
    \caption{Selected samples of \ours{} on Blocked Gibbs sampling.}
    \label{fig:gibbs_examples}
\end{figure}

\begin{figure}[H]
    \centering
    \includegraphics[width=\linewidth]{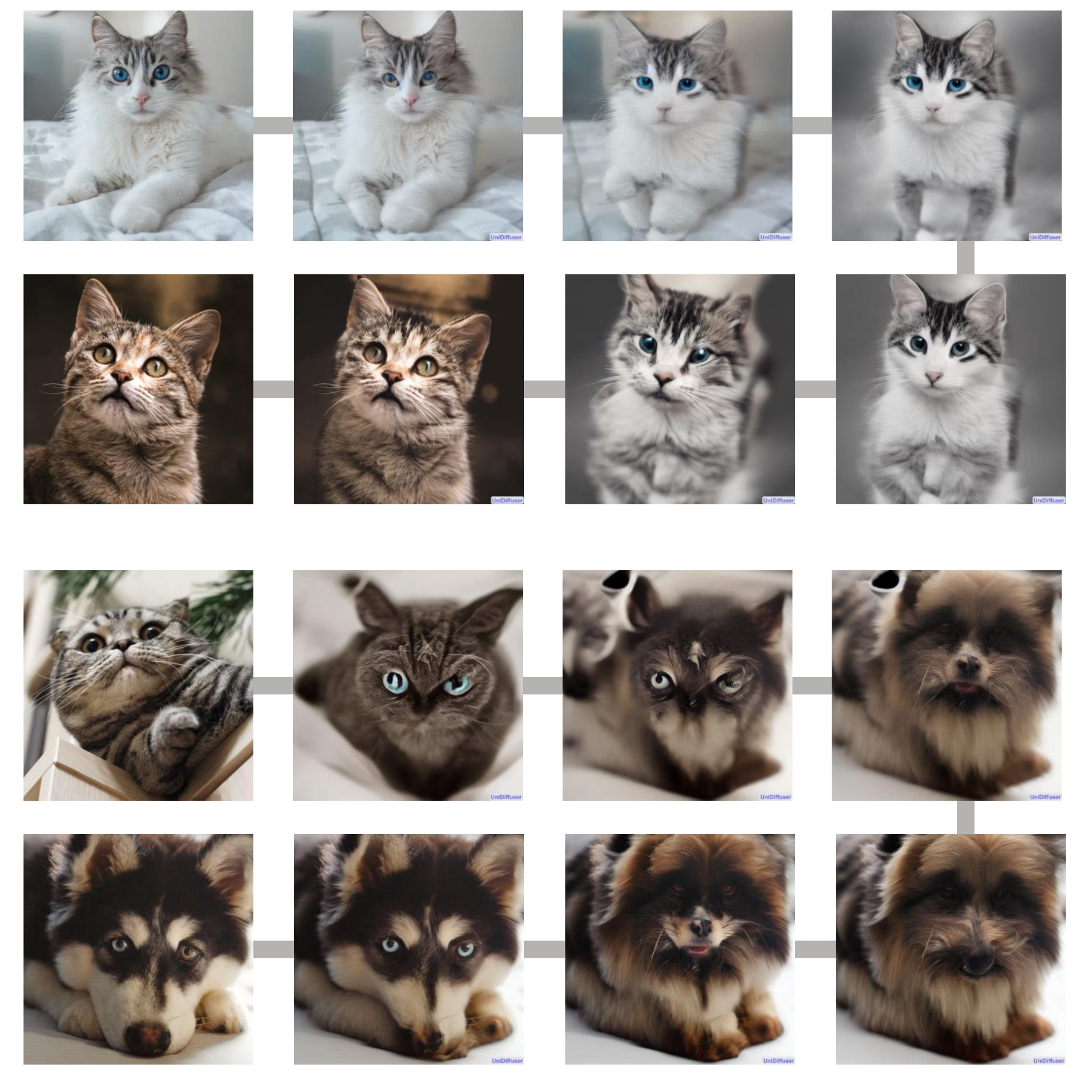}
    \caption{Selected samples of \ours{} on interpolation of two images in the wild.}
    \label{fig:interpolation_examples}
\end{figure}

Below we present more samples of \ours{} on joint generation (Figure~\ref{fig:joint_examples}), text-to-image generation (Figure~\ref{fig:t2i_examples}), image-to-text generation (Figure~\ref{fig:i2t_examples}), unconditional image generation (Figure~\ref{fig:i_examples}), unconditional text generation (Figure~\ref{fig:t_examples}), image variation (Figure~\ref{fig:i2t2i_examples}), text variation (Figure~\ref{fig:t2i2t_examples}), blocked Gibbs sampling (Figure~\ref{fig:gibbs_examples}) and interpolation (Figure~\ref{fig:interpolation_examples}).

\section{The Training and Sampling Algorithms}
\label{sec:alg}

In Algorithm~\ref{alg:train}, we present the training algorithm of \ours{}.
In Algorithm~{\ref{alg:uncond_sample_x},\ref{alg:cond_sample_xy},\ref{alg:joint_sample}}, we present all sampling procedure of \ours{} by taking the DDPM sampler~\cite{ho2020denoising} as an example. Note that any other learning-free efficient sampler, such as DDIM~\cite{song2020denoising}, Analytic-DPM~\cite{bao2022analytic} and DPM-Solver~\cite{lu2022dpm,lu2022dpmpp}, is directly applicable.

\begin{figure}[H]
\begin{minipage}[t]{0.48\textwidth}
\begin{algorithm}[H]
    \caption{Training}
    \label{alg:train}
    \begin{algorithmic}[1]
        \STATE \textbf{repeat}
        \STATE\ \  $\vx_0, \vy_0 \sim q(\vx_0, \vy_0)$
        \STATE\ \  $t^x, t^y \sim \mathrm{Uniform}(\{1,2,\dots,T\})$
        \STATE\ \  $\vepsilon^x, \vepsilon^y \sim \gN(\vzero, \mI)$
        \STATE\ \  Let $\vx_{t^x} = \sqrt{\overline{\alpha}_{t^x}} \vx_0 + \sqrt{1 - \overline{\alpha}_{t^x}} \vepsilon^x$
        \STATE\ \  Let $\vy_{t^y} = \sqrt{\overline{\alpha}_{t^y}} \vy_0 + \sqrt{1 - \overline{\alpha}_{t^y}} \vepsilon^y$
        \STATE\ \  Take gradient step on \\ \quad \ \  $\nabla_\vtheta \|\vepsilon_\vtheta(\vx_{t^x},\vy_{t^y}, t^x,t^y) - [\vepsilon^x,\vepsilon^y] \|_2^2$
        \STATE \textbf{until} converged
    \end{algorithmic}
\end{algorithm}
\end{minipage}
\hspace{.1cm}
\begin{minipage}[t]{0.48\textwidth}
\begin{algorithm}[H]
    \caption{Unconditional sampling of $\vx_0$ \\ (similar for unconditional sampling of $\vy_0$)}
    \label{alg:uncond_sample_x}
    \begin{algorithmic}[1]
        \STATE $\vx_T \sim \gN(\vzero, \mI)$
        \STATE \textbf{for} $t=T,\dots,1$ \textbf{do}
        \STATE\ \ $\vz^x \sim \gN(\vzero, \mI)$ if $t > 1$, else $\vz^x = \vzero$
        \STATE\ \ $\vepsilon \sim \gN(\vzero, \mI)$
        \STATE\ \ $\vx_{t-1} = \frac{1}{\sqrt{\alpha_t}} \left(\vx_t - \frac{\beta_t}{\sqrt{1 - \overline{\alpha}_t}} \vepsilon_\vtheta^x(\vx_t, \vepsilon, t, T) \right) + \sigma_t \vz^x$
        \STATE \textbf{end for}
        \STATE \textbf{return} $\vx_0$
    \end{algorithmic}
\end{algorithm}
\end{minipage}

\begin{minipage}[t]{0.48\textwidth}
\begin{algorithm}[H]
    \caption{Sampling of $\vx_0$ conditioned on $\vy_0$ \\ (similar for sampling of $\vy_0$ conditioned on $\vx_0$)}
    \label{alg:cond_sample_xy}
    \begin{algorithmic}[1]
        \STATE $\vx_T \sim \gN(\vzero, \mI)$
        \STATE \textbf{for} $t=T,\dots,1$ \textbf{do}
        \STATE\ \ $\vz^x \sim \gN(\vzero, \mI)$ if $t > 1$, else $\vz^x = \vzero$
        \STATE\ \ $\vx_{t-1} = \frac{1}{\sqrt{\alpha_t}} \left(\vx_t - \frac{\beta_t}{\sqrt{1 - \overline{\alpha}_t}} \vepsilon_\vtheta^x(\vx_t, \vy_0, t, 0) \right) + \sigma_t \vz^x$
        \STATE \textbf{end for}
        \STATE \textbf{return} $\vx_0$
    \end{algorithmic}
\end{algorithm}
\end{minipage}
\hspace{.1cm}
\begin{minipage}[t]{0.48\textwidth}
\begin{algorithm}[H]
    \caption{Joint sampling of $\vx_0, \vy_0$}
    \label{alg:joint_sample}
    \begin{algorithmic}[1]
        \STATE $\vx_T, \vy_T \sim \gN(\vzero, \mI)$
        \STATE \textbf{for} $t=T,\dots,1$ \textbf{do}
        \STATE\ \ $\vz^x, \vz^y \sim \gN(\vzero, \mI)$ if $t > 1$, else $\vz^x, \vz^y = \vzero$
        \STATE\ \ $\vx_{t-1} = \frac{1}{\sqrt{\alpha_t}} \left(\vx_t - \frac{\beta_t}{\sqrt{1 - \overline{\alpha}_t}} \vepsilon_\vtheta^x(\vx_t, \vy_t, t, t) \right) + \sigma_t \vz^x$
        \STATE\ \ $\vy_{t-1} = \frac{1}{\sqrt{\alpha_t}} \left(\vy_t - \frac{\beta_t}{\sqrt{1 - \overline{\alpha}_t}} \vepsilon_\vtheta^y(\vx_t, \vy_t, t, t) \right) + \sigma_t \vz^y$
        \STATE \textbf{end for}
        \STATE \textbf{return} $\vx_0, \vy_0$
    \end{algorithmic}
\end{algorithm}
\end{minipage}
\end{figure}

\section{Summary of Classifier-Free Guidance Models}
\label{sec:summary_cfg}

As mentioned in Section~\ref{sec:cfg_free}, \ours{} can perform classifier-free guidance for free for conditional and joint generation. We summarize models for classifier-free guidance in Table~\ref{tab:models}. We also include the models for unconditional sampling.

\begin{table}[H]
    \caption{\textbf{Models for different sampling tasks.} $\vepsilon^x$ and $\vepsilon^y$ are sampled from the standard Gaussian noise. $s$ is the classifier-free guidance scale.}
    \label{tab:models}
    \vskip 0.1in
    \centering
    \begin{tabular}{cc}
    \toprule
        Task & Model \\
    \midrule
        Joint sampling & $\hat{\vepsilon}_\vtheta(\vx_t, \vy_t, t) = (1+s) \vepsilon_\vtheta(\vx_t, \vy_t, t, t) - s [\vepsilon_\vtheta^x(\vx_t, \vepsilon^y, t, T), \vepsilon_\vtheta^y(\vepsilon^x, \vy_t, T, t)]$ \\
    \arrayrulecolor{black!30}\midrule
        Sample $\vx_0$ conditioned on $\vy_0$ & $\hat{\vepsilon}_\vtheta^x(\vx_t, \vy_0, t) = (1+s)\vepsilon_\vtheta^x(\vx_t, \vy_0, t, 0) -s \vepsilon_\vtheta^x(\vx_t, \vepsilon^y, t, T)$ \\
    \arrayrulecolor{black!30}\midrule
        Sample $\vy_0$ conditioned on $\vx_0$ & $\hat{\vepsilon}_\vtheta^y(\vy_t, \vx_0, t) = (1+s)\vepsilon_\vtheta^y(\vx_0, \vy_t, 0, t) -s \vepsilon_\vtheta^y(\vepsilon^x, \vy_t, T, t)$ \\
    \arrayrulecolor{black!30}\midrule
        Unconditional sampling of $\vx_0$ & $\hat{\vepsilon}_\vtheta^x(\vx_t, t) = \vepsilon_\vtheta^x(\vx_t, \vepsilon^y, t, T)$ \\
    \arrayrulecolor{black!30}\midrule
        Unconditional sampling of $\vy$ & $\hat{\vepsilon}_\vtheta^y(\vy_t, t) = \vepsilon_\vtheta^y(\vepsilon^x, \vy_t, T, t)$ \\
    \arrayrulecolor{black}\bottomrule
    \end{tabular}
\end{table}

\section{Details of the U-ViT}
\label{sec:detail_uvit}

We train a U-ViT for the joint noise prediction network, whose detailed configuration is presented in Table~\ref{tab:uvit_cfg}.

\begin{table}[H]
    \caption{The configuration of U-ViT.}
    \label{tab:uvit_cfg}
    \vskip 0.1in
    \centering
    \begin{tabular}{cccccc}
    \toprule
        Patch size & \#Layers & Hidden size & MLP size & \#Heads & \#Params \\
    \midrule
        2 & 31 & 1536 & 6144 & 24 & 952M \\
    \bottomrule
    \end{tabular}
\end{table}

\section{Details of the GPT-2 Text Decoders}
\label{sec:gpt2}

\begin{figure}[H]
    \centering
    \includegraphics[width=0.5\linewidth]{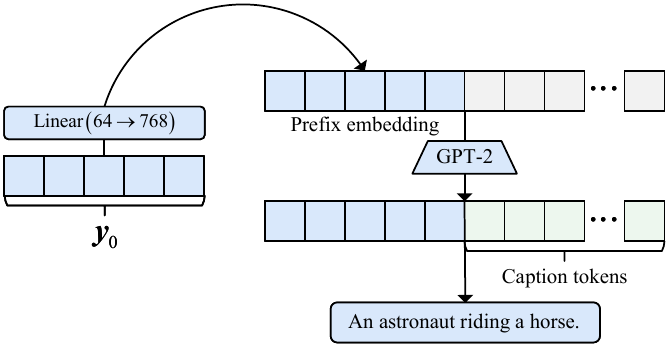}
    \caption{\textbf{GPT-2 Text Decoder.} $\vy_0$ is fed into GPT-2 as a prefix embedding~\cite{mokady2021clipcap}, and reconstructs the text autoregressively based on the information of $\vy_0$}
    \label{fig:gpt2}
\end{figure}

As mentioned in Section~\ref{sec:encode}, we firstly encode the text $\mathbf{T}$ into a low-dimensional embedding $\vy_0$ through a CLIP and a linear layer, i.e., $\vy_0 = \mathrm{Linear}(\mathrm{CLIP}(\mathbf{T}))$, and then reconstruct the original text using the GPT-2 by feeding $\vy_0$ as a prefix embedding. The reconstruction pipeline of the GPT-2 decoder is illustrated in Figure~\ref{fig:gpt2}. We finetune the GPT-2 and the linear layer with the following autoregressive loss:
\begin{align*}
    \min_\vphi \E_{\mathbf{T}} [\log p(\mathbf{T} | \vy_0)] = \E_{\mathbf{T}} [\sum_{i=1}^N \log p(\mathbf{T}_i | \mathbf{T}_{1:i-1}, \vy_0)],
\end{align*}
where $\vphi$ denotes the parameters of the linear layer and the GPT-2.

We finetune the 124M parameter GPT-2 text decoder on texts of LAION-2B-en dataset \cite{schuhmann2022laion}, which contains 2.3B image-text pairs. Following ClipCap~\cite{mokady2021clipcap}, we use the AdamW optimizer~\cite{loshchilov2017decoupled} with a learning rate of 2e-5 and 5K warm-up steps. We train the decoder with 235K steps using a batch size of 768. When generating texts, we use the beam search strategy with a beam size of 5 and a maximum length of 67.

\begin{figure}[H]
    \centering
    \includegraphics[width=\linewidth]{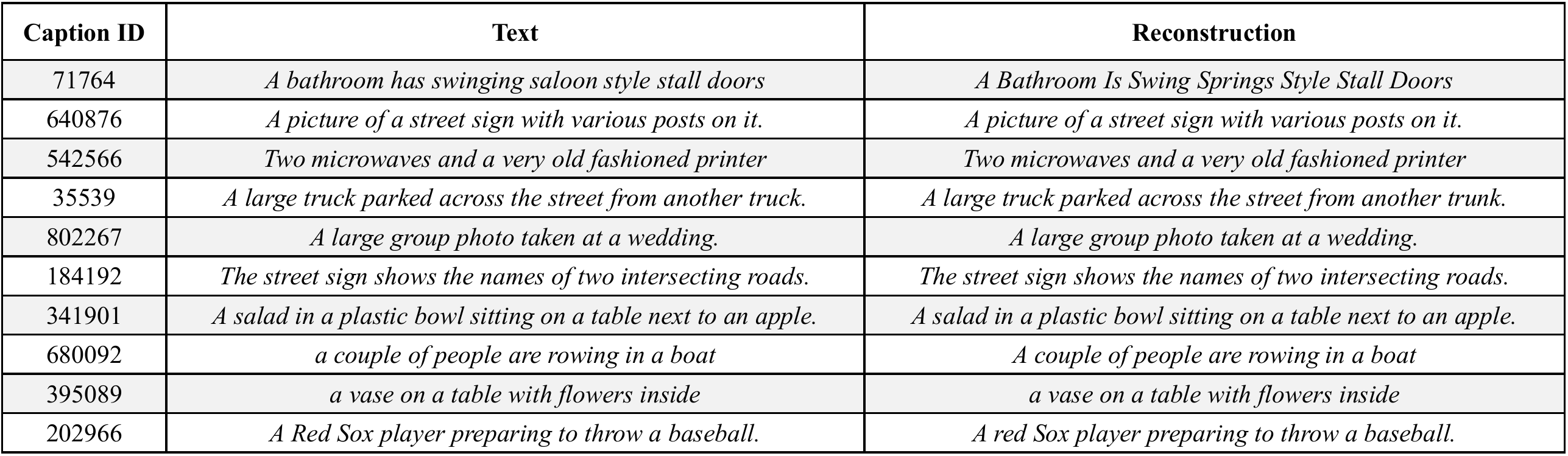}
    \caption{\textbf{Text reconstruction examples.} The GPT-2 text decoder reconstructs the text well.}
    \label{fig:text_reconstruction}
\end{figure}

In our experiments, the embedding dimension of $\vy_0$ is set to 64, and it still reconstructs the text well. Indeed, we get a BLEU-1~\cite{papineni2002bleu} score of 0.969 and a BLEU-4 score of 0.894 between reference texts and reconstructed texts on MS-COCO test set (Karpathy split~\cite{karpathy2015deep}). We present some examples in Figure~\ref{fig:text_reconstruction}, where the input texts are reconstructed very well.

% we reconstruct 5,000 captions from the MS-COCO test set according to the Karpathy split~\cite{karpathy2015deep}

% \fan{show Bleu[1…4]? Bleu1: 0.969, Bleu2: 0.943, Bleu3: 0.918, Bleu4: 0.894}  

\section{The Interpolation Algorithm}
\label{sec:interpolation_alg}

We present the formalized procedure of interpolating two images using \ours{} in Algorithm~\ref{alg:interpolate}.

\begin{algorithm}[H]
    \caption{Interpolate two images $\mathbf{I}^a$ and $\mathbf{I}^b$}
    \label{alg:interpolate}
    \begin{algorithmic}[1]
        \STATE \textbf{Input:} the spherical interpolation parameter $\theta \in [0, 1]$ ($\theta=0$ leads to $\mathbf{I}^a$ and $\theta=1$ leads to $\mathbf{I}^b$)
        \STATE Let $\hat{\vepsilon}_\vtheta^y(\vy_t, \vx_0, t) = (1+s)\vepsilon_\vtheta^y(\vx_0, \vy_t, 0, t) -s \vepsilon_\vtheta^y(\vepsilon^x, \vy_t, T, t)$ be the image-to-text model
        \STATE Let $\hat{\vepsilon}_\vtheta^x(\vx_t, \vy_0, t) = (1+s)\vepsilon_\vtheta^x(\vx_t, \vy_0, t, 0) -s \vepsilon_\vtheta^x(\vx_t, \vepsilon^y, t, T)$ be the text-to-image model
        \STATE Encode $\mathbf{I}^a$ and $\mathbf{I}^b$ to get their latent embeddings $\vx_0^a$ and $\vx_0^b$
        \STATE $\vy_T \sim \gN(\vzero, \mI)$
        \STATE $\vy_0^a = \mathrm{DPM}\text{-}\mathrm{Solver}( \mathrm{initial\_state} = \vy_T, \mathrm{start\_time} = T, \mathrm{end\_time} = 0, \mathrm{model} = \hat{\vepsilon}_\vtheta^y(\vy_t, \vx_0^a, t) )$
        \STATE $\vy_0^b = \mathrm{DPM}\text{-}\mathrm{Solver}( \mathrm{initial\_state} = \vy_T, \mathrm{start\_time} = T, \mathrm{end\_time} = 0, \mathrm{model} = \hat{\vepsilon}_\vtheta^y(\vy_t, \vx_0^b, t) )$
        \STATE $\vx_T^a = \mathrm{DPM}\text{-}\mathrm{Solver}( \mathrm{initial\_state} = \vx_0^a, \mathrm{start\_time} = 0, \mathrm{end\_time} = T, \mathrm{model} = \hat{\vepsilon}_\vtheta^x(\vx_t, \vy_0^a, t) )$
        \STATE $\vx_T^b = \mathrm{DPM}\text{-}\mathrm{Solver}( \mathrm{initial\_state} = \vx_0^b, \mathrm{start\_time} = 0, \mathrm{end\_time} = T, \mathrm{model} = \hat{\vepsilon}_\vtheta^x(\vx_t, \vy_0^b, t) )$
        \STATE $\vy_0^\theta = \mathrm{slerp}(\vy_0^a, \vy_0^b, \theta)$
        \STATE $\vx_T^\theta = \mathrm{slerp}(\vx_T^a, \vx_T^b, \theta)$
        \STATE $\vx_0^\theta = \mathrm{DPM}\text{-}\mathrm{Solver}( \mathrm{initial\_state} = \vx_T^\theta, \mathrm{start\_time} = T, \mathrm{end\_time} = 0, \mathrm{model} = \hat{\vepsilon}_\vtheta^x(\vx_t, \vy_0^\theta, t) )$
        \STATE Decode $\vx_0^\theta$ to get the image $\mathbf{I}^\theta$
        \STATE \textbf{Return} $\mathbf{I}^\theta$
    \end{algorithmic}
\end{algorithm}

\clearpage
\section{Comparison of Examples}
\label{sec:cmp_example}

\begin{figure}[H]
    \centering
    \includegraphics[width=0.8\columnwidth]{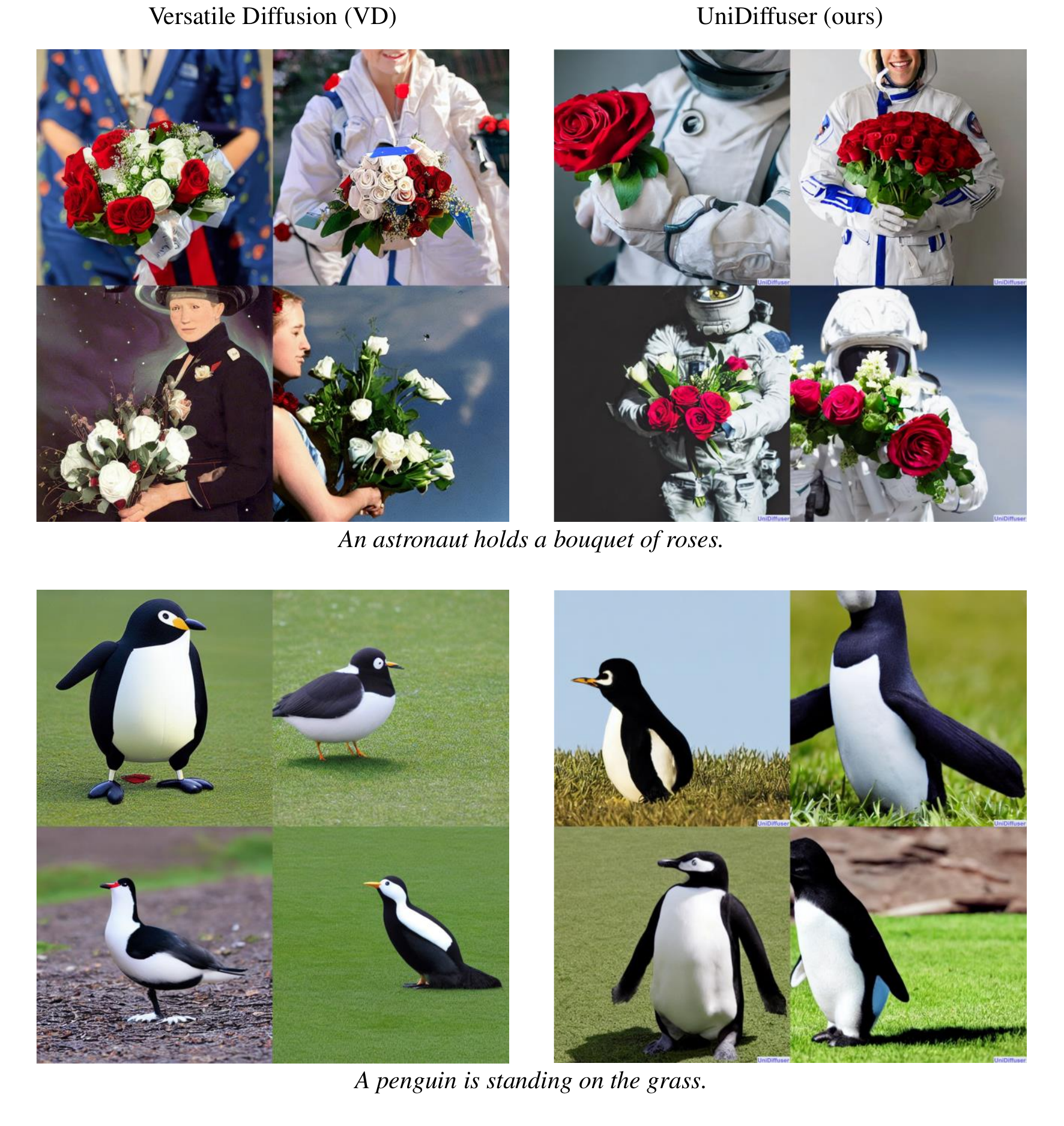}\vspace{-.6cm}
    \caption{Comparison of examples between \ours{} and VD on text-to-image generation. Samples generated by \ours{} align the texts better than VD.}
    \label{fig:compare_t2i_app1}
\end{figure}

\begin{figure}[H]
    \centering
    \includegraphics[width=0.8\columnwidth]{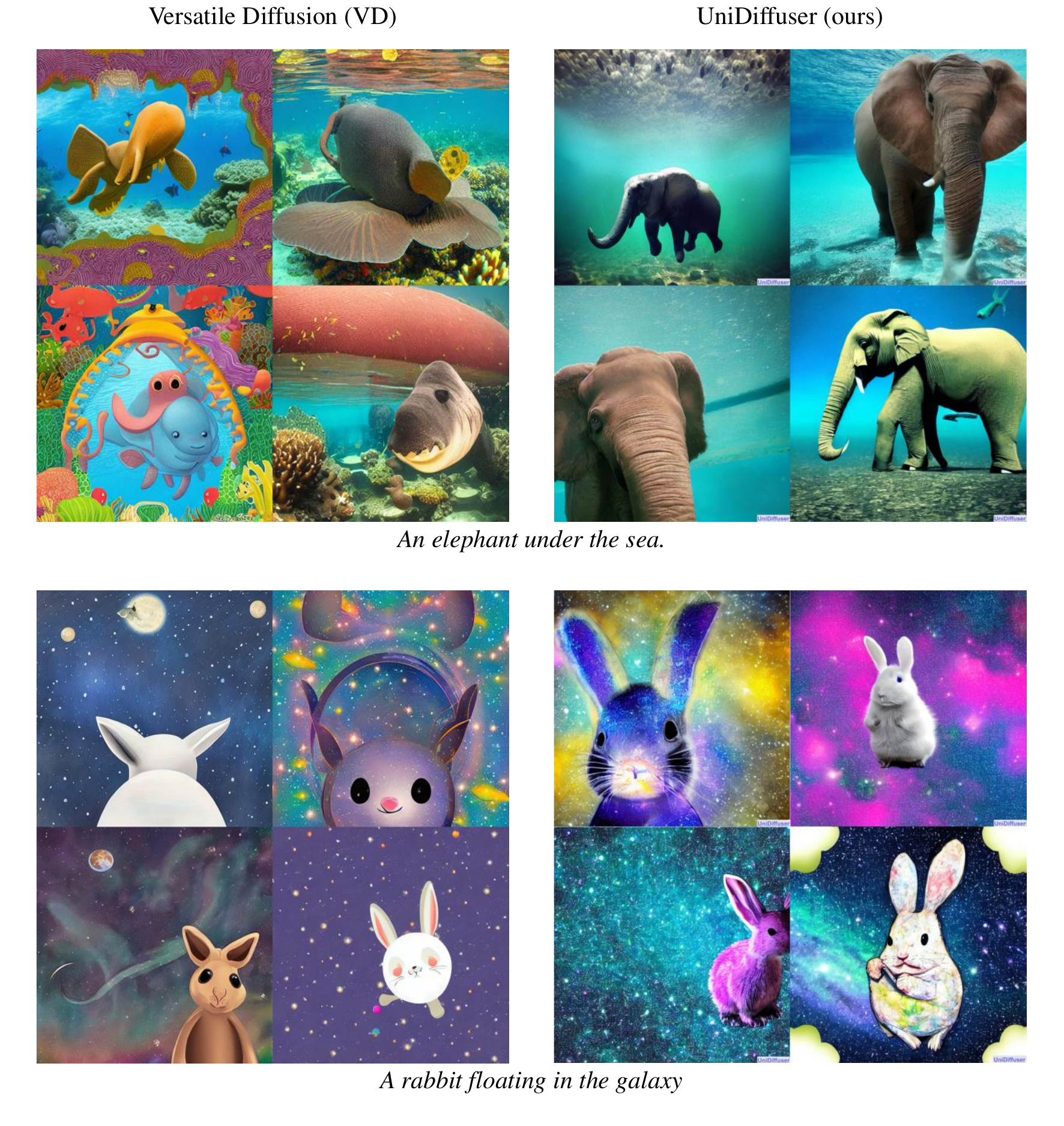}\vspace{-.6cm}
    \caption{Comparison of examples between \ours{} and VD on text-to-image generation. Samples generated by \ours{} align the texts better than VD.}
    \label{fig:compare_t2i_app2}
\end{figure}

\begin{figure}[H]
    \centering
    \includegraphics[width=0.8\columnwidth]{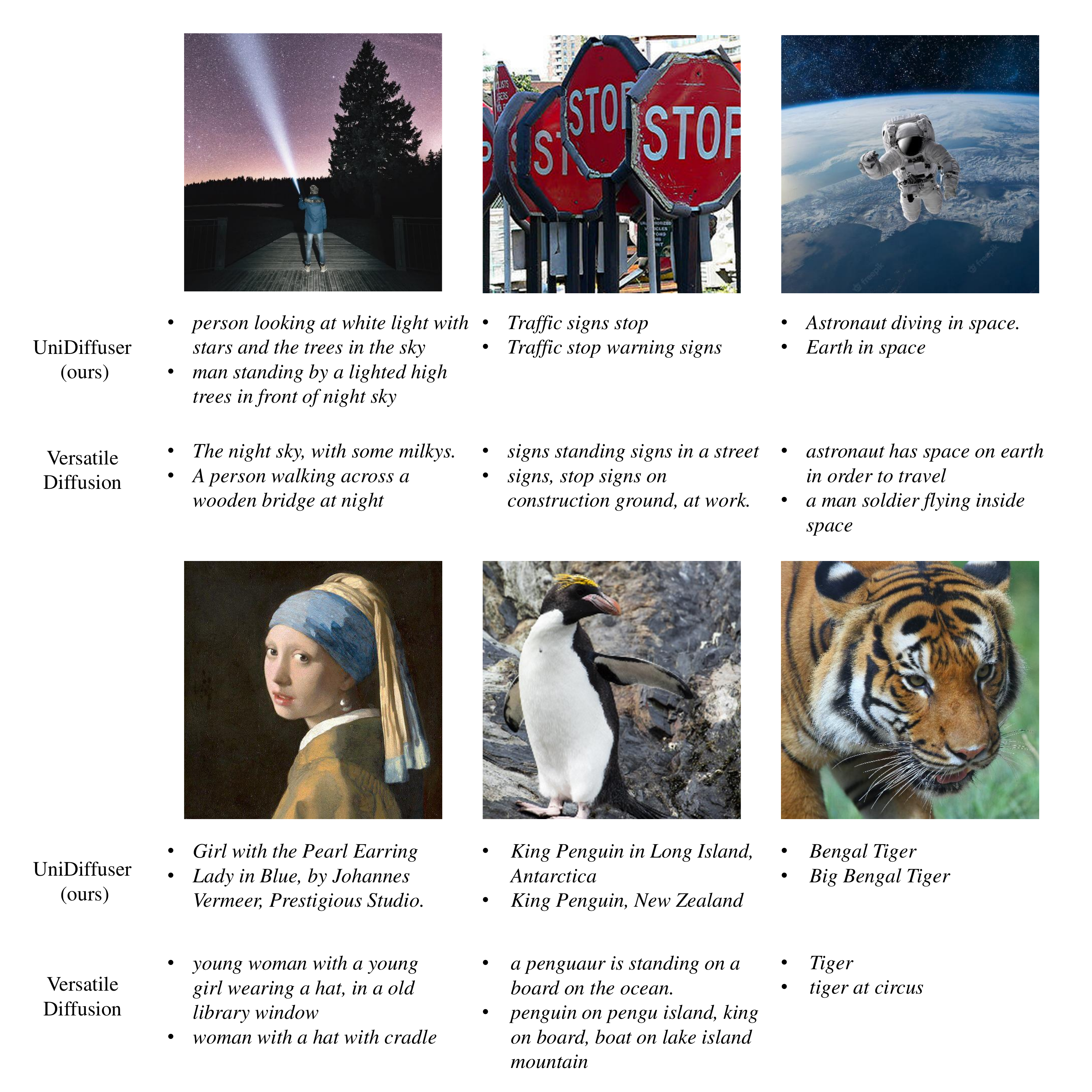}\vspace{-.6cm}
    \caption{Comparison of examples between \ours{} and VD on image-to-text generation. Samples generated by \ours{} align the images better than VD.}
    \label{fig:compare_i2t}
\end{figure}

In Figure~\ref{fig:compare_t2i_app1} and Figure~\ref{fig:compare_t2i_app2}, we present more examples on text-to-image generation of \ours{} and Versatile Diffusion. Samples generated by \ours{} align better with the texts than VD. In Figure~\ref{fig:compare_i2t}, we present examples on image-to-text generation of \ours{} and Versatile Diffusion. Samples generated by \ours{} align better with the images than VD.

\section{Efficiency Comparison}

In Table~\ref{tab:eff}, we compare the model size, inference time and memory (for generating 10 samples with 25 denoising steps on one A100 80GB GPU), and the training cost of UniDiffuser with other bespoke and general-purpose models. Results of other methods are obtained according to the official code or paper when they are available.

UniDiffuser is more efficient than both Stable Diffusion and Versatile Diffusion in terms of inference time and memory. Besides, compared to Stable Diffusion, UniDiffuser introduces only 10\% extra parameters to support five tasks (i.e., image, text, text-to-image, image-to-text, and image-text pair generation) with comparable training cost. Compared to the general-purpose model Versatile Diffusion, UniDiffuser has fewer parameters, while achieving superior results (as presented in the main paper).

\begin{table}[H]
    \caption{Comparison of model size and computational cost.}\label{tab:eff}
    \vskip 0.1in
    \centering
    \scalebox{0.85}{
    \begin{tabular}{lllll}
    \toprule
        & Model size & Inference time & Inference memory & Training cost \\
    \midrule
    \emph{Bespoken models} \\
    ~~~~DALL$\cdot$E 2~\cite{ramesh2022hierarchical} & 4.5B & -- & -- & -- \\
    ~~~~Imagen~\cite{saharia2022photorealistic} & 2B & -- & -- & -- \\
    ~~~~Parti~\cite{yu2022scaling} & 20B & -- & -- & -- \\
    ~~~~Stable Diffusion$^\dagger$~\cite{rombach2022high} & 860M & 25.43s & 67.83GB & 150K (A100 40GB GPU hours) \\
    \midrule
    \emph{General-purpose models} \\
    ~~~~Versatile Diffusion$^\dagger$~\cite{xu2022versatile} & 2566M & 23.89s & 76.53GB & -- \\
    ~~~~\ours{} (\textbf{ours}) & 952M & 19.77s & 48.30GB & 59K (A100 80GB GPU hours) \\
    \bottomrule
    \end{tabular}}
    \vspace{-.3cm}
\end{table}

\section{Licences}

Datasets:
\begin{itemize}
    \item LAION-5B~\cite{schuhmann2022laion}: Creative Common CC-BY 4.0 license
    \item MS-COCO~\cite{lin2014microsoft}: Creative Commons Attribution 4.0 License
\end{itemize}

Pretrained models:
\begin{itemize}
    \item GPT-2~\cite{radford2019language}: MIT License
    \item CLIP~\cite{radford2021learning}: MIT License
    \item Image autoencoder from Stable Diffusion~\cite{rombach2022high}: CreativeML Open RAIL-M License
\end{itemize}

\end{document}